# From fragmented data to actionable design: Physics-calibrated learning for plastic upcycling

Jingyang Bai [a, †], Zijia Wang [b, †], Xiangyi Long [a], Marcos Millan [c, d], Binjian Nie [e, *], and Mingyue Ding [f, g, *]

[a] Department of Chemical Engineering, Imperial College London, South Kensington Campus, London SW7 2AZ, UK

[b] Department of Electrical and Electronic Engineering, Imperial College London, South Kensington Campus, London SW7 2AZ, UK

[c] Interdisciplinary Research Center for Refining and Advanced Chemicals, King Fahd University of Petroleum and Minerals, Dhahran 31261, Saudi Arabia

[d] Department of Chemical Engineering, King Fahd University of Petroleum & Minerals, Dhahran 31261, Saudi Arabia

[e] Department of Engineering Science, The University of Oxford, Parks Road, OX3 1PJ, Oxford, UK

[f] School of Power and Mechanical Engineering, Wuhan University, Wuhan, 430072, China

[g] Academy of Advanced Interdisciplinary Studies, Wuhan University, Wuhan, 430072, China

[†] These authors contributed equally to this work.

* Corresponding author: binjian.nie@eng.ox.ac.uk; dingmy@whu.edu.cn

**ABSTRACT**

Thermochemical upgrading of plastic waste is a key upcycling pathway, yet the experimental literature is fragmented by heterogeneous conditions and incomplete reporting. Complete-case learning would retain only 10.99% of the curated experiments, while target imputation can introduce biased supervision. Here we develop a Physics-Calibrated, Missingness-Gated, and Load-Balanced Mixture-of-Experts (PC-MG-MoE) framework that converts structured missingness into an informative learning signal. PC-MG-MoE learns directly from partially observed experiments without target imputation, reconstructs physically consistent product distributions, accommodates cross-laboratory heterogeneity, and provides interpretable model behaviour rather than black-box prediction alone. Under stringent source-grouped validation, it achieved the lowest aggregate absolute error among the evaluated models, supporting engineering screening under cross-laboratory heterogeneity. Wet-lab experiments provide an external comparison, showing key composition-dependent trends. Implemented as an interactive web-based workflow, PC-MG-MoE enables forward screening, physics-grounded constrained inverse design, targeted experimental planning that supports reduced experimental workload and trial-and-error, and laboratory-specific adaptation with new platform-specific data. This work establishes a transferable framework for converting fragmented literature data into experimentally actionable guidance for model-guided plastic upcycling and broader thermochemical systems.

# 1. Introduction

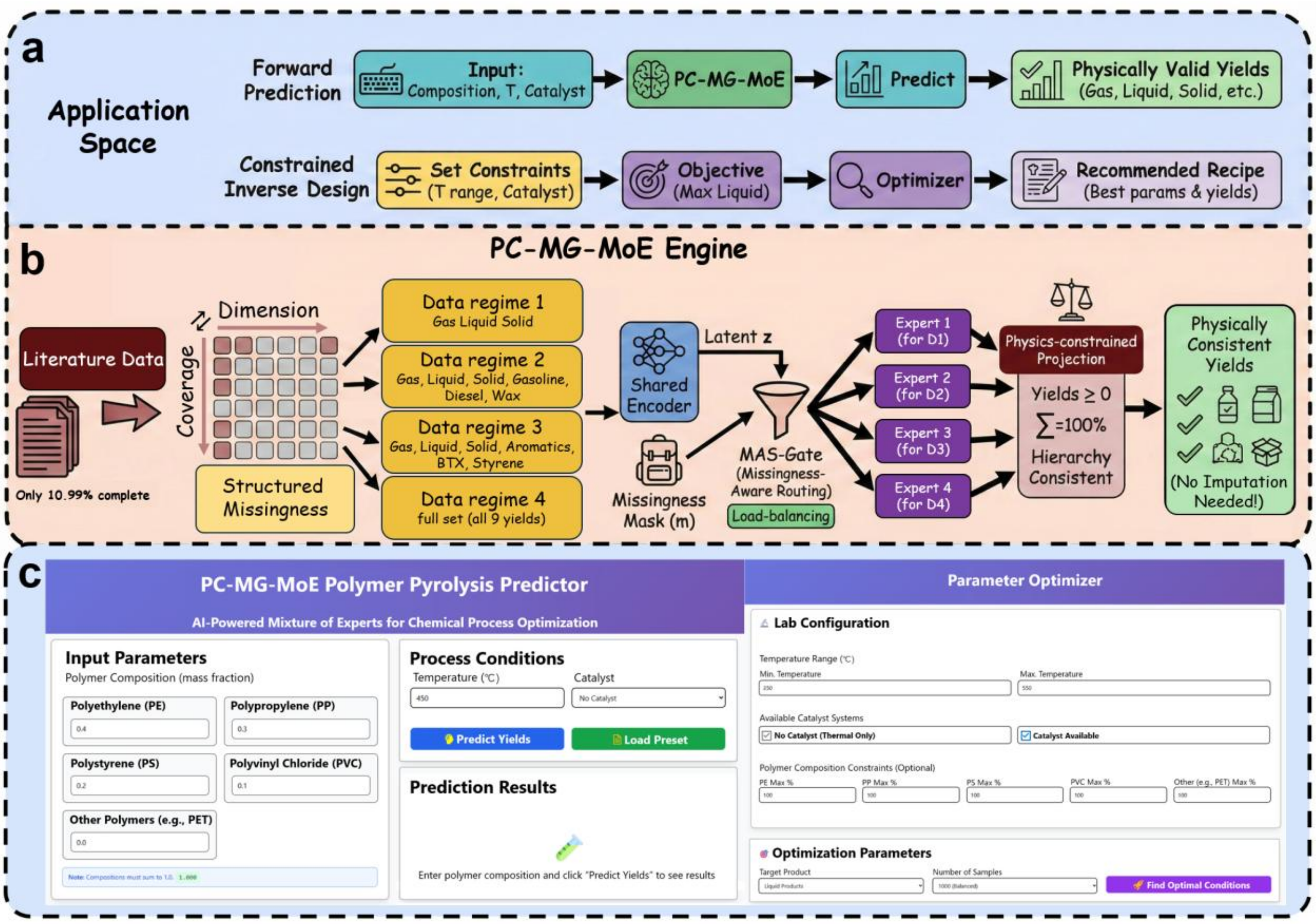

**Figure 1. Deployment, application workflow and architecture of the PC-MG-MoE framework.** (a) Application workflow for forward prediction and constrained inverse design, linking user-defined inputs to physically valid yield predictions and recommended reaction conditions. (b) Internal architecture of the PC-MG-MoE engine, showing structured-missingness representation, shared encoding, missingness-aware expert routing, regime-specialised experts and physics calibration for enforcing non-negativity, mass balance and hierarchical consistency across multi-level product yields. (c) Simplified schematic of the deployed web-based interfaces, including the Polymer Pyrolysis Predictor for forward yield prediction and the Parameter Optimizer for constrained inverse design under user-defined compositional and operational constraints.

Plastic products play an essential role in modern society due to their favourable mechanical and chemical properties, yet their rapidly increasing production has exacerbated a global waste crisis [1-3]. Despite their high embedded chemical and energy value, less than 10% of plastic waste is effectively recycled, while most is incinerated, landfilled, or released into the environment, causing persistent contamination, greenhouse gas emissions, and substantial resource losses [4-7]. Developing effective routes to valorise plastic waste is therefore both an environmental necessity and a critical challenge in sustainable chemical engineering.

Compared with conventional physical recycling routes that often suffer from property degradation and low-value products [8-11], chemical recycling has emerged as a promising strategy for plastic upcycling. Among chemical recycling technologies, thermochemical upgrading routes, particularly plastic pyrolysis, have attracted significant attention because they can convert waste polymers into fuels, chemicals, and carbonaceous materials under relatively flexible operating conditions [12-14], with the potential to reduce greenhouse gas emissions relative to incineration [15,16]. However, fully realising the potential of pyrolysis requires systematic optimisation across a high-dimensional design space involving feedstock composition, reaction temperature, and catalyst selection [17,18].

Despite extensive experimental work, translating literature data into actionable guidance remains difficult. Reported experiments vary widely in feed composition, temperature and catalytic systems, reactor configuration and heating conditions, and crucially, studies often report only a subset of product yields, such as total liquid yield without a full breakdown into gasoline-range, diesel-range and wax fractions, or aromatic subfamilies. Systematic exploration by one-factor-at-a-time experimentation is costly, and many studies probe only narrow local regions, leaving the broader response surface sparsely sampled.

Recent work has begun to apply machine learning (ML) to polymer pyrolysis and catalytic upgrading, including neural-network and tree-based models that map operating conditions to yields or product families. Most existing studies adopt a straightforward strategy of aggregating reported experiments into a single dataset for supervised learning. Although these approaches have advanced yield prediction, they often remain limited by target imputation or sample exclusion, weakly account for laboratory heterogeneity, provide limited support for constrained inverse design, and rarely enforce physical feasibility.

Among the limitations of this aggregated-table paradigm, structured missingness in the target space is particularly important. Conventional supervised learning pipelines either discard incomplete samples, reducing already limited data, or impute missing targets, introducing pseudo-labels that can bias supervision and produce physically inconsistent predictions. This limitation is evident in recent ML studies on plastic pyrolysis. For instance, Belden et al. [19] had to remove several variables and applied K-nearest neighbour (KNN) imputation, explicitly acknowledging that incomplete reporting contributed to model uncertainty. Similarly, Xu et al. [20] substituted missing entries with median or average values, while Li et al. [21] highlighted the difficulty of extracting consistent information across publications due to inconsistent reporting formats. As more target variables are considered, the number of fully observed samples decreases sharply, creating a trade-off between dataset dimensionality and model reliability, especially when missingness correlates with chemistry, measurement scope or publication practice.

More fundamentally, prevailing strategies implicitly treat incompletely reported data as unusable or secondary, leaving a large fraction of legacy literature data underexploited for quantitative analysis and experimental design. As a result, many experimentally investigated conditions cannot be meaningfully compared or reused across studies, limiting their value for systematic reaction analysis. Therefore, there is a clear need for modelling approaches specifically designed to cope with the sparse, heterogeneous, and incompletely reported pyrolysis datasets.

Beyond incomplete reporting, a deeper limitation lies in how existing studies treat heterogeneity across laboratories. In literature-derived datasets, missingness follows structured, laboratory-dependent observation patterns rather than random omissions. Researchers typically report yields aligned with their analytical scope, reactor configuration, and catalyst systems, leading different laboratories to consistently report distinct subsets of targets.

If these reporting patterns are treated as noise, important information about how the data were generated is discarded. Simply aggregating all literature data into a single table and treating unreported targets as missing values overlooks this laboratory heterogeneity. Such naive aggregation implicitly assumes that experiments conducted in different laboratories follow comparable experimental and measurement regimes, which is rarely the case for catalytic pyrolysis studies. As a result, models trained on aggregated datasets may learn spurious correlations or introduce systematic bias when learning from partially reported outcomes.

Finally, beyond data-related challenges, the fundamental limitation of existing data-driven studies lies in their limited relevance to experimental decision-making in chemical engineering practice. Most approaches emphasise model benchmarking and predictive metrics, often with limited model-behaviour interpretation, while offering little support for screening feasible operating conditions or prioritising experiments under physical and operational constraints. As a result, conventional models primarily function as post hoc predictors, often reproducing intuitive yet trivial trends (for example, polystyrene-rich conditions favouring light-oil production in pyrolysis systems [22]), but they provide limited guidance for process-oriented decision-making or inverse design.

Collectively, existing studies have not yet addressed these engineering limitations within an integrated and deployment-ready framework.

To model such heterogeneous and systematically incomplete experimental regimes, mixture-of-experts (MoE) models provide a natural mechanism for conditional computation, where different experts can specialise on different data regimes. However, standard MoE routing typically relies on input features alone and does not explicitly exploit the observation pattern itself, which is a key signal in datasets where missingness reflects systematic measurement scope.

To address these limitations, we introduce a Physics-Calibrated, Missingness-Gated, and Load-Balanced Mixture-of-Experts (PC-MG-MoE) framework that bridges data-driven modelling with practical experimental decision-making in plastic pyrolysis. Rather than discarding incompletely reported experiments or artificially imputing missing targets, the framework explicitly represents laboratory-specific reporting regimes through an observation mask and uses mask-aware expert routing to exploit partially observed data within a unified model.

In doing so, the framework transforms structured missingness from a limitation into an information source, enabling legacy experimental data that would otherwise be excluded to be leveraged for quantitative analysis. Importantly, this formulation allows laboratory heterogeneity to be explicitly modelled rather than implicitly ignored, providing a unified learning architecture that remains consistent across diverse experimental regimes.

Beyond predictive modelling, the framework is designed to support practical use under realistic engineering constraints. By enforcing physically feasible yield relationships, the proposed approach produces chemically coherent predictions and enables both forward prediction and constrained inverse design, allowing feasible operating conditions to be screened and prioritised prior to experimentation. Permutation-importance and controlled-response analyses are further used to examine whether the trained model relies on chemically meaningful descriptors rather than acting as a black-box predictor. To examine its predictive capability under laboratory conditions, we compare PC-MG-MoE outputs with wet-lab pyrolysis measurements obtained from mixed-plastic experiments. This shifts the framework from passive yield prediction towards actionable, process-oriented decision support under realistic experimental constraints.

The trained model is further packaged into an interactive web application supporting forward prediction and constrained inverse design, making the framework directly accessible for exploratory and decision-support workflows. This deployment structure also provides a route for laboratory-specific adaptation as new data are generated from individual experimental platforms. An overview of the model architecture, application workflow, and deployment interface is provided in Figure 1.

Taken together, the main contribution of this work lies not only in predicting plastic pyrolysis yields, but in establishing a missingness-aware and physics-calibrated learning framework that converts fragmented, partially reported thermochemical literature data into physically feasible, interpretable and experimentally actionable design guidance. By integrating source-grouped evaluation, model-behaviour analysis, an interactive web-based workflow and constrained inverse design, the framework moves beyond black-box retrospective yield prediction towards model-guided experimental planning.

To our knowledge, this work is among the first frameworks to systematically leverage incompletely reported literature datasets for practical experimental decision-making in plastic upcycling. The code and web-workflow implementation are publicly available to facilitate reuse and further laboratory-specific adaptation.

## 2. Methods

### 2.1. Dataset Compilation and Curation

We first extracted 713 plastic-pyrolysis experiments from 108 literature sources. Each record contains seven input descriptors, including feedstock composition (polyethylene (PE), polypropylene (PP), polystyrene (PS), polyvinyl chloride (PVC) and Other, e.g., polyethylene terephthalate (PET); wt.%), reaction temperature (°C) and a binary catalyst flag (1 = catalyst present, 0 = non-catalytic). The target vector comprises nine yields reported on a consistent mass basis (wt.% of feed), including gas, liquid and solid yields; gasoline-range hydrocarbons (C5–C12, hereafter gasoline), diesel-range hydrocarbons (C13–C20, hereafter diesel) and wax (>C21) yields; and aromatic subfamilies (total aromatics, benzene–toluene–xylenes (BTX) and styrene yields). To construct a consistent modelling dataset, we harmonised reported product yields to a consistent mass basis (wt.% of feed), standardised polymer fractions to sum to 100 wt.%, and removed duplicated or ambiguous entries. Targets were screened using automated consistency and plausibility checks (non-negativity, gas+liquid+solid ≤ 100 wt.%, subfraction sums and hierarchy constraints). Entries lacking core inputs or failing these checks were excluded from model training, yielding the curated dataset comprising 282 experiments from 73 sources. The full extraction (713 entries from 108 sources) is publicly available in the accompanying data package.

## 2.2 ML Model Design: Physics-Calibrated, Missingness-Gated, and Load-Balanced Mixture-of-Experts (PC-MG-MoE)

Predicting product distributions from literature-derived pyrolysis datasets is challenging because the data are both heterogeneous and pervasively incomplete. Different studies report different subsets of outputs, and key quantities such as gasoline, diesel, wax, aromatics, BTX and styrene are missing in most experiments. Classical regression approaches either discard partially observed samples or impute missing targets, at the cost of reduced data utilisation or physically inconsistent labels. To address these limitations, we design a Physics-Calibrated, Missingness-Gated, and Load-Balanced Mixture-of-Experts (PC-MG-MoE) architecture that leverages the structure of the dataset while adhering to chemical laws, with the detailed architecture shown in Figure S1.

We consider a literature-derived pyrolysis dataset composed of $N$ experiments $\{(x_k, y_k^{obs})\}_{k=1}^{N}$. The input vector $x_k$ aggregates feedstock descriptors and process conditions. The target $y_k \in \mathbb{R}^T$ represents the product-yield distribution over a fixed set of $T = 9$ outputs ordered as Gas, Liquid, Solid, Gasoline, Diesel, Wax, Aromatics, BTX, Styrene.

Due to heterogeneous reporting scopes across studies, only a subset of targets is available for each experiment. We encode this structured missingness with a binary mask $m_k \in \{0,1\}^T$, where $(m_k)_t = 1$ indicates that target $t$ is reported and $(m_k)_t = 0$ otherwise. The partially observed label vector is denoted $y_k^{obs}$, and supervision is applied only on reported targets via $m_k$ (no target imputation is performed).

As summarised in Figure S1, PC-MG-MoE integrates a shared representation, missingness-aware routing, regime-specialised experts and physics calibration. The observation mask is used by the Missingness-Aware Softmax Gating (MAS-Gate) network to condition expert allocation on both process descriptors and reporting scope. The resulting mixed prediction is subsequently calibrated using a physics-calibration operator, implemented as a projection operator, to enforce non-negativity, macro-product mass balance, liquid-cut consistency and aromatic-hierarchy constraints.

Detailed mathematical definitions of the gating mechanism, expert aggregation, physics calibration, masked regression loss and regularisation terms are provided in Section S1 of the Supporting Information.

**2.3 Evaluation Protocol**

The model is evaluated using source-grouped five-fold cross-validation, where experiments originating from the same literature source are held out together. Because sparsely reported targets can produce unstable fold-level $R^2$ when a held-out fold contains few observations or low target variance, we additionally constructed a target-balanced source-grouped sensitivity split that still assigns whole sources to folds but balances non-missing target counts and target sums. Leakage checks confirmed zero source-ID and experiment-signature overlap between training and test folds in this sensitivity split. The source-grouped evaluation protocol is substantially more stringent than random experiment-level splitting because it prevents records from the same laboratory, reactor setup or reporting style from appearing simultaneously in training and test folds. It therefore provides a more realistic assessment of cross-source generalisation. Continuous input and output variables were min–max scaled before model training, with the same scaling procedure applied consistently across all models.

Performance metrics include mean squared error (MSE), mean absolute error (MAE), and coefficient of determination ($R^2$). MAE is interpreted as the primary engineering error measure, whereas $R^2$ is used as a stricter variance-explained diagnostic. Negative $R^2$ values are reported without truncation and indicate limited variance-explained cross-source generalisation. Metrics are computed only on targets observed in the test fold and then macro-averaged across targets. Results are reported as mean ± standard deviation over five folds. For sparse targets such as BTX and styrene, conservative fallbacks are additionally included in a hybrid evaluation setting to avoid extrapolation outside observed regimes. Per-target metrics are also reported to illustrate how predictive performance varies with label density and reporting regime.

Baseline models include Random Forest [23], Gradient Boosting and its variants (CatBoost and XGBoost) [24-26], multilayer perceptron (MLP) neural networks [27], Ridge regression [28], and Support vector regression (SVR) [29] baselines. Missing targets were handled using the following strategies: mean imputation for Random Forest, CatBoost, Gradient Boosting, XGBoost, Ridge, and one MLP baseline; chemistry-aware filling for SVR; and masked-loss training for the masked-loss MLP baseline. All baselines share the same input features and are evaluated under the same source-grouped split protocol.

### 2.4 Laboratory-Scale Wet Experiments for Model–Experiment Comparison

To assess whether the trained model could reproduce experimentally measured product distributions, laboratory-scale wet experiments were conducted using a controlled plastic pyrolysis reactor system. The experimental conditions were converted into the same input descriptors used by the model, and the resulting model-predicted outputs were compared with the experimentally measured product yields. These wet-lab data served as an external model–experiment comparison and were not used for model training, hyperparameter optimisation, or baseline benchmarking. The wet experiments were conducted in a plastic pyrolysis reactor, and detailed information on the reactor flow sheet, product collection and analytical procedures is provided in Section S2 of the Supporting Information.

## 3. Results and Discussion

### 3.1 Structured Missingness Regimes

Target missingness in literature-reported plastic pyrolysis is strongly structured. In the curated dataset, liquid yield is reported for all 282 experiments, whereas gas and solid yields are each missing in 8.16% of cases (Figure 2). Higher-value subfractions exhibit substantially higher missingness. Gasoline, diesel and wax are each missing in 69.15% of experiments, BTX and styrene are each missing in 65.96%, and total aromatics is missing in 41.49%. Only 31 experiments (10.99%) are fully observed, illustrating the severe data-retention penalty imposed by complete-case analysis.

These omissions are not random; rather, they systematically co-occur in recurrent reporting regimes, as shown by the binary observation masks in Figure 2(b). The full set of reporting regimes and their corresponding target combinations are summarised in Table S1. For modelling, these raw observation patterns are further grouped into four regime families (D1–D4), which define the expert structure shown in Figure S1. This pattern reflects common analytical pipelines rather than random dropout, indicating that target missingness is linked to how different studies quantify pyrolysis products.

This structure has direct consequences for modelling strategy (Figure 2(c)). A complete-case approach that keeps all nine targets retains only 31 experiments (10.99%), whereas dropping the high-missing targets retains all 282 experiments but collapses the output space to macro yields. PC-MG-MoE avoids this either–or trade-off by learning from all partially observed targets while using the observation mask to guide expert routing, enabling full-vector prediction at inference time despite incomplete supervision during training.

More broadly, these characteristics expose a fundamental limitation of conventional regression models, which typically require complete observations and therefore rely on either discarding incomplete samples or imputing missing values. This creates an inherent trade-off between retaining dataset size and preserving output dimensionality, motivating modelling strategies that can learn directly from sparse and partially observed pyrolysis datasets without sacrificing data utilisation or output resolution.

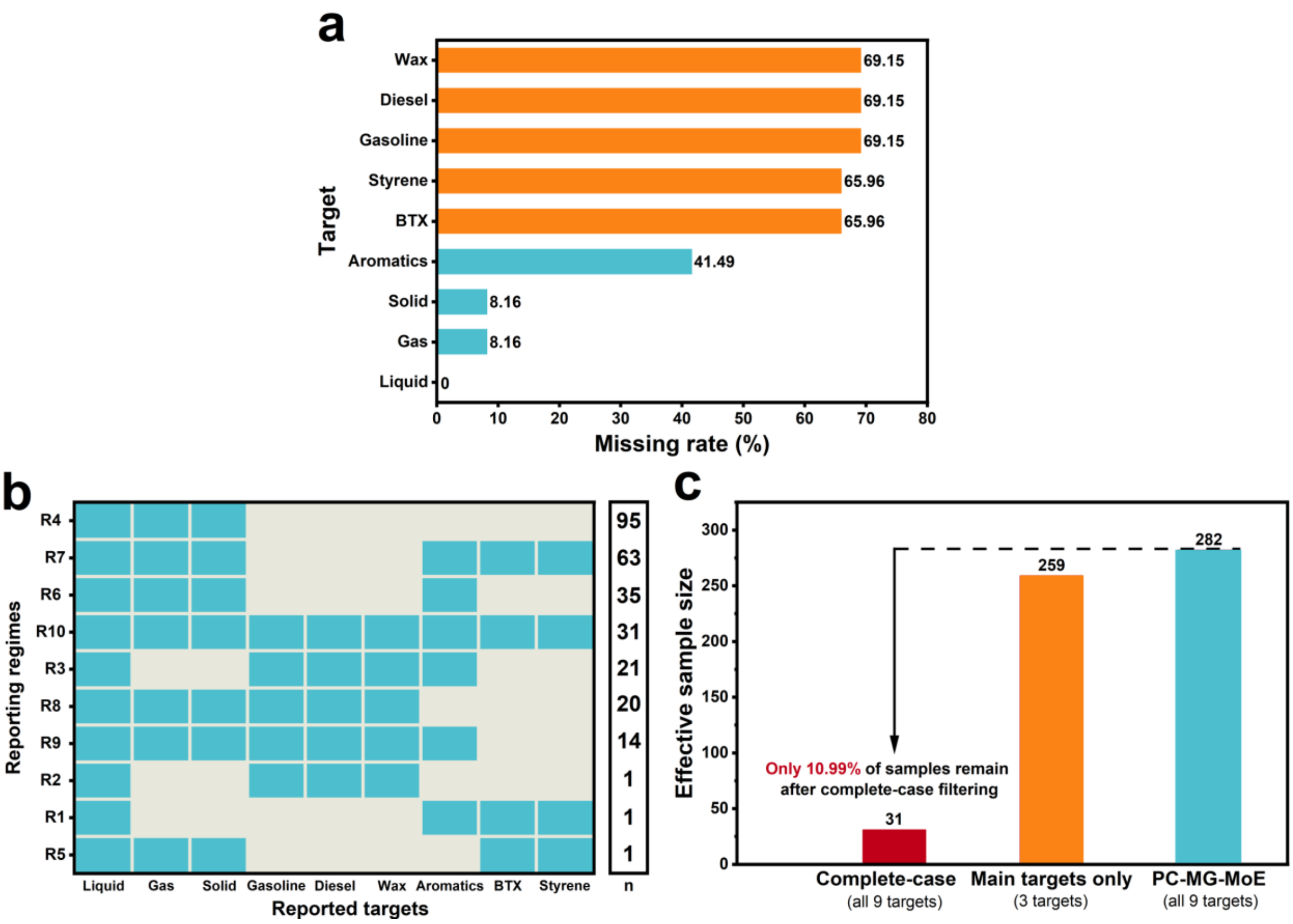

**Figure 2. Structured target missingness in plastic pyrolysis datasets and its impact on modelling.** (a) Missing rates across targets, showing substantially higher missingness for subfractions; (b) Binary observation masks revealing recurrent reporting regimes; (c) Effective sample size under different strategies, highlighting the trade-off between data retention and output dimensionality, and how PC-MG-MoE utilises all samples while preserving the full target space.

### 3.2 Baseline Comparison

To benchmark the performance of PC-MG-MoE, we compare against a set of representative baseline models covering tree-based, neural, linear, and kernel regressors (see Section 2.3 for details). All baseline models were trained and evaluated under the source-grouped evaluation protocol, where experiments from the same literature source are held out together to assess cross-laboratory generalisation (Figure 3 and Table 1).

Under the source-grouped evaluation protocol, PC-MG-MoE achieves the best aggregate MAE (lowest error) among all compared models (13.98 ± 3.77 wt.%), improving over random forests (14.68 ± 4.71 wt.%) and gradient boosting (15.40 ± 4.59 wt.%). Because MAE is expressed directly in wt.% yield, it provides a practically interpretable measure of absolute prediction error for engineering screening and experiment prioritisation. The corresponding aggregate $R^2$ remains negative for all methods, indicating that variance-explained performance is limited under source-grouped cross-laboratory generalisation when target labels are sparse and experimental protocols vary across literature sources. The $R^2$ results followed the same relative trend, with PC-MG-MoE giving the least negative aggregate value among the evaluated models.

This performance gap is particularly informative under the source-grouped split, which is intentionally stringent because entire sources are held out, introducing distribution shifts in polymer feeds, catalysts, and analytical pipelines. As a result, aggregate $R^2$ is negative for many baselines, reflecting that predictive relationships learned from one source do not necessarily transfer directly to another. In this setting, the lower MAE achieved by PC-MG-MoE is particularly relevant for engineering use, as it indicates reduced absolute yield error even when variance-explained metrics remain limited by source heterogeneity and sparse supervision. A target-balanced source-grouped sensitivity analysis reduced the aggregate $R^2$ penalty from −0.292 to −0.115 for the hybrid PC-MG-MoE evaluation, and an independent nested tabular sweep achieved a similar value (−0.111; MAE 13.80 wt.%). These sensitivity results indicate that the most severe negative $R^2$ values are amplified by sparse-target fold imbalance, while showing that $R^2$ remains a stringent diagnostic under cross-source evaluation. Detailed per-target sensitivity results are provided in Section S3 of the Supporting Information.

Per-target analysis reveals where the model generalises and where the data remain limiting (Figure S3 and Table S2). Targets with broad coverage (gas and liquid) achieve positive mean $R^2$ (0.27 and 0.18, respectively). Total aromatics also shows weakly positive $R^2$ (0.10), whereas sparsely reported targets such as BTX and styrene (96 labels each) exhibit negative $R^2$ across folds. These trends mirror the effective supervision available for each target and motivate modelling strategies that share information across regimes without discarding incomplete samples. In the target-balanced sensitivity split, gas, liquid, aromatics and styrene improved to positive or near-zero $R^2$ for PC-MG-MoE, and the nested tabular sweep produced positive $R^2$ for BTX. However, gasoline-range products remained more challenging, underscoring that high-value subfractions are not uniformly recoverable from the currently available cross-laboratory metadata.

Beyond pointwise predictive error, the ability to produce a complete, physically feasible yield vector is critical for downstream use. PC-MG-MoE produces full nine-target predictions for every query, enabling scenario screening and inverse design even when a given laboratory historically reports only a subset of targets.

Importantly, when trained on the merged dataset, most conventional baselines relied on explicit missing-target handling strategies (mean imputation or chemistry-aware filling), while one neural baseline used masked-loss training. In contrast, PC-MG-MoE handles partial supervision natively through missingness-aware gating and masked supervision. The observed performance gains therefore cannot be attributed to imputation choices but instead reflect the advantage of explicitly modelling structured missingness and expert specialisation.

These results demonstrate that native handling of incomplete output spaces, combined with physics calibration and balanced expert routing, yields a modest but consistent improvement while enabling physically consistent full-vector prediction without requiring target imputation.

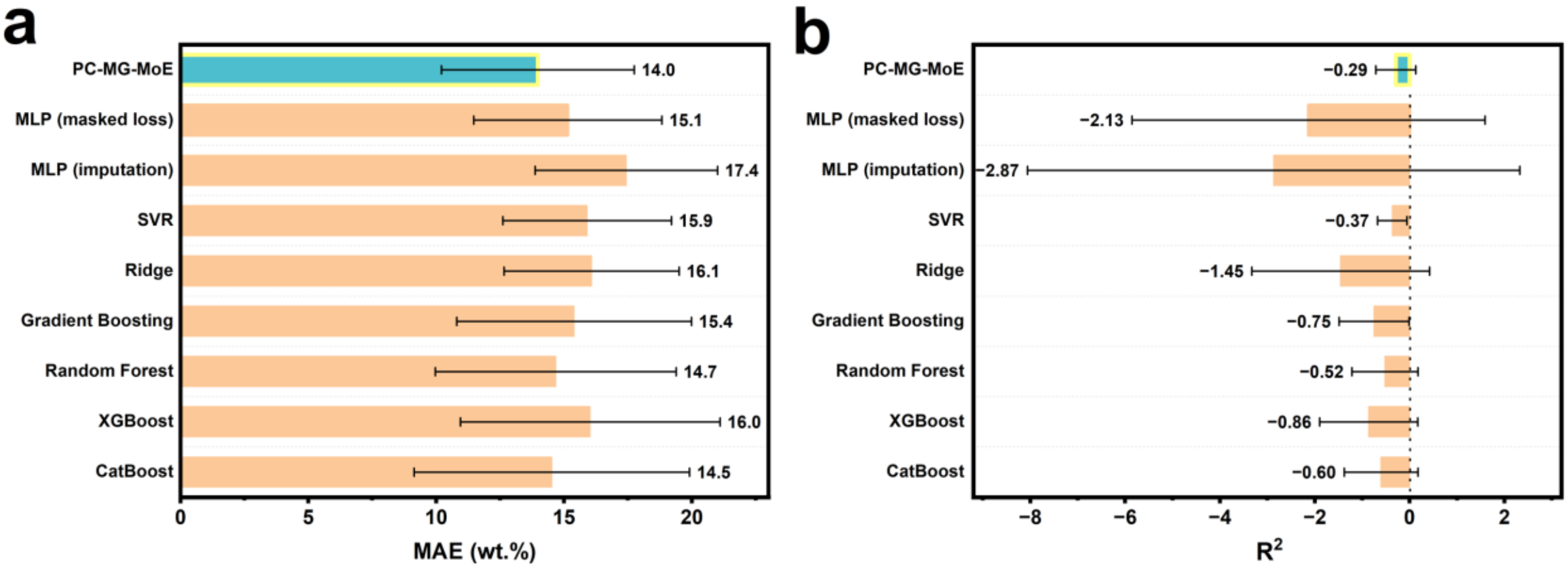

**Figure 3. Baseline comparison under source-grouped evaluation.** (a) Mean absolute error (MAE) across baseline models and PC-MG-MoE, reported in wt.% yield. (b) Coefficient of determination ($R^2$) across the same models. Error bars indicate variability across source-grouped cross-validation folds.

**Table 1.** Performance comparison of PC-MG-MoE against baseline regression models under structured missingness.

| Model | MSE↓ | MAE↓ | $R^2$↑ | Missing Strategy |
|---|---|---|---|---|
| **PC-MG-MoE (Ours)** | **416.8 ± 180.7** | **13.98 ± 3.77** | **−0.29 ± 0.42** | **Native handling** |
| Random Forest | 478.9 ± 287.0 | 14.68 ± 4.71 | −0.52 ± 0.70 | Mean imputation |
| CatBoost | 464.0 ± 313.1 | 14.52 ± 5.08 | −0.60 ± 0.78 | Mean imputation |
| MLP (masked loss) | 455.3 ± 173.3 | 15.14 ± 3.68 | −2.13 ± 3.72 | Masked loss |
| Gradient Boosting | 533.2 ± 299.3 | 15.40 ± 4.59 | −0.75 ± 0.74 | Mean imputation |
| SVR | 553.0 ± 188.5 | 15.91 ± 3.30 | −0.37 ± 0.31 | Chemistry-aware |
| XGBoost | 585.8 ± 382.1 | 16.02 ± 5.38 | −0.86 ± 1.03 | Mean imputation |
| Ridge | 480.3 ± 194.4 | 16.08 ± 3.42 | −1.45 ± 1.87 | Mean imputation |
| MLP (imputation) | 538.3 ± 182.6 | 17.44 ± 3.57 | −2.87 ± 5.19 | Mean imputation |

### 3.3 Ablation Experiments

To quantify the contribution of individual architectural components in PC-MG-MoE, a series of ablation studies were conducted to examine expert specialisation, routing mechanisms and physics-based constraints.

We therefore separate two ablation questions. First, source-grouped component ablations quantify whether each architectural choice remains useful under cross-source generalisation. In this strict setting, the full pure PC-MG-MoE configuration gives an aggregate MAE of 14.68 ± 3.55 wt.% and $R^2$ of −0.56 ± 0.48, while removing mask information from the gate gives similar aggregate error and hard routing degrades performance (MAE 15.53 ± 3.42 wt.%; $R^2$ −0.86 ± 1.05; Table S3). These results indicate that aggregate $R^2$ under source-held-out validation is dominated by domain shift and sparse labels, whereas mask-aware soft routing primarily improves robustness and feasibility rather than producing a large global $R^2$ gain. Second, a within-source architecture diagnostic compares a single expert and MoE under a non-source- grouped split (Figure S4(a) and Table S4). These within-source results are used only to compare architectural behaviour under matched data conditions, while cross-laboratory generalisation is evaluated using the source-grouped ablations described above. Beyond expert specialisation, the routing mechanism also plays a key role. A routing ablation further highlights the importance of soft, mask-aware routing. Replacing soft routing with mask-based hard routing degrades performance, consistent with the need to share information across regimes while still specialising experts. Hard routing is particularly detrimental in this setting because reporting regimes overlap in targets. For example, some studies report aromatics without BTX/styrene, whereas others report the full aromatic breakdown. Soft routing allows the model to share information across such partially overlapping supervision while still permitting regime-dependent specialisation.

Physics-based constraints play a critical role in the architecture (Figure S4(b)). The constraint ablation shows that removing mass-balance, non-negativity, and component-consistency constraints yields only a marginal change in validation MSE (329.9 vs 335.0, <2% difference). We retain constraints for physical plausibility at negligible predictive cost.

At the prediction stage, the physics projection substantially improves feasibility diagnostics. Before projection, predicted yield vectors frequently violate basic constraints (average violation rates across folds: non-negativity 45%, mass-balance bound 100%, and liquid-cut consistency 99%). After projection, violations drop to 0% for non-negativity, mass balance and liquid-cut constraints, and the rate of aromatic-hierarchy violations is also reduced (from 0.43 to 0.30; Figure S4(c)). Projection adjustments can be substantial (Figure S4(d)), underscoring that feasibility checks are a necessary complement to predictive metrics when deploying surrogate models for engineering design.

Together, these results establish a clear module-level logic for PC-MG-MoE, in which expert specialisation improves learning over a single multi-output model, mask-aware soft routing improves information sharing across partially overlapping reporting regimes, and physics calibration ensures physically feasible predictions at negligible predictive cost. This supports the need to combine these components within a unified architecture for heterogeneous and partially observed reaction systems.

### 3.4 Expert Routing and Model-Behaviour Analysis

The behaviour of a MoE model fundamentally depends on whether the gating network can distribute samples across experts in a meaningful and interpretable manner. The learned gate adapts routing probabilities to the observation regimes. Overall, expert utilisation remains balanced across folds, while regime-specific preferences emerge in the routing matrix (Figure 4), indicating that the model leverages the missingness mask as an informative conditioning signal rather than treating missing targets as noise.

Across all folds, the average routing weights for the four experts are 0.22, 0.23, 0.34 and 0.21, respectively, indicating that multiple experts are actively utilised in practice rather than collapsing to a single predictor (Figure 4(a)). Routing also varies systematically with reporting regime (Figure 4(b)). For example, the macro-only regime (gas–liquid–solid) shows increased weight on an expert specialised in macro-yield relationships, whereas regimes that include gasoline/diesel/wax cuts shift probability mass towards a different expert.

This regime-conditioned behaviour supports the interpretation that the observation mask captures experimental context (analytical scope and measurement focus) that is not explicit in the input features alone. This combination of balanced capacity usage and regime-specific specialisation enables the MoE to effectively utilise its representational capacity while maintaining robustness across heterogeneous observation regimes.

To further address the black-box nature often associated with ML models, the behaviour of the trained PC-MG-MoE was examined using permutation feature importance and controlled-response analysis (Figures 4(c) and S5). Unlike simple dataset-level correlation analysis, permutation importance probes model behaviour directly by randomly permuting each input descriptor after training and quantifying the resulting deterioration in output-specific prediction performance. The model assigned high importance to PS content for aromatics and styrene, while temperature and catalyst flag contributed strongly to bulk product distributions and liquid-fraction outputs (Figure 4(c)).

Controlled-response analysis, performed by varying one model input while keeping the remaining inputs fixed, further showed that increasing PS content increased predicted aromatics and styrene while decreasing wax, and that temperature changes induced distinct shifts in gas, liquid and aromatic-related outputs (Figure S5). These patterns are consistent with established pyrolysis knowledge, suggesting that the model captures chemically meaningful input–output relationships rather than relying only on spurious correlations in the training data.

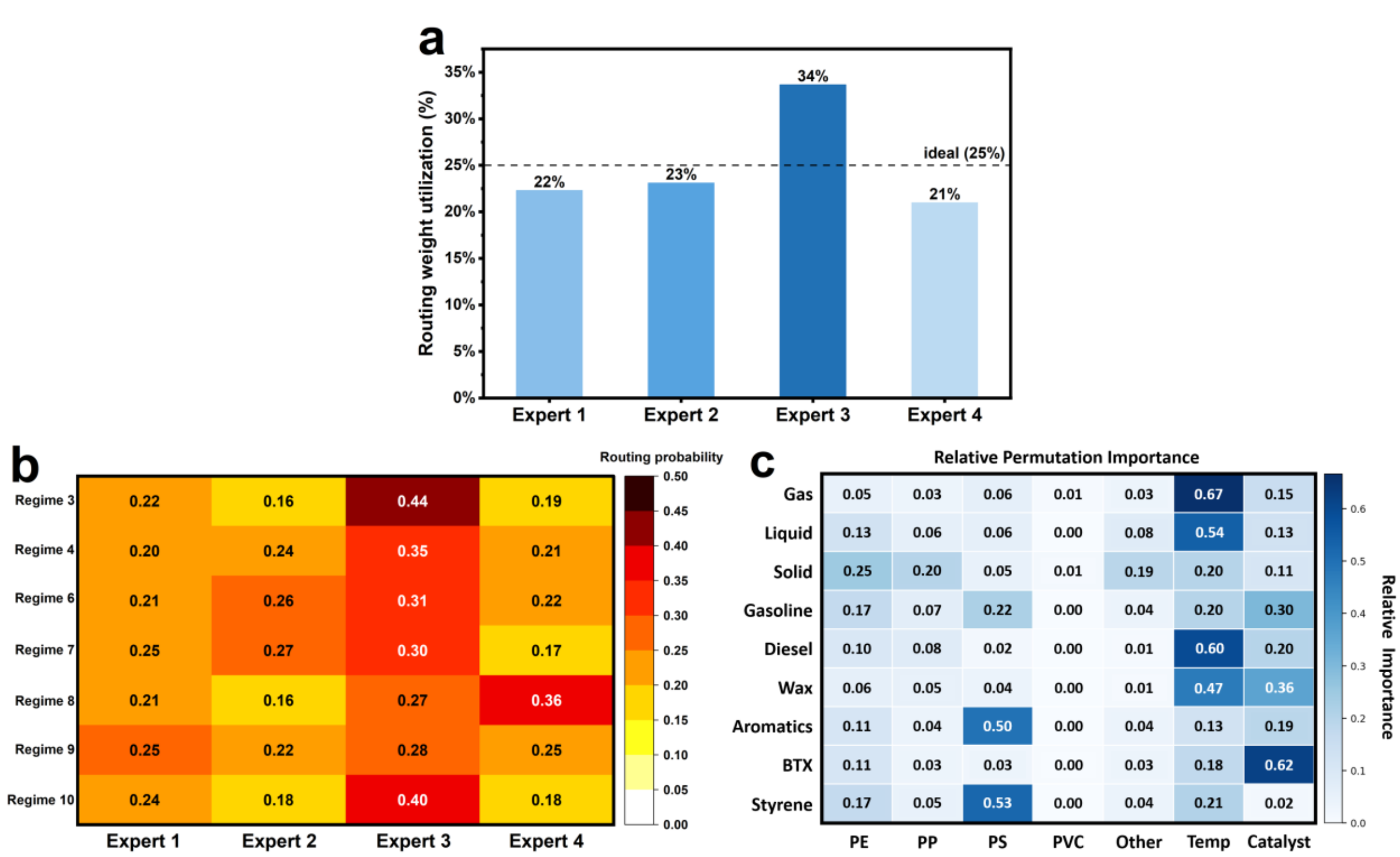

**Figure 4. Expert routing behaviour and model-behaviour analysis of the PC-MG-MoE model.** (a) Overall expert utilisation; (b) Routing heatmap showing average expert weights across reporting regimes. (c) Permutation feature importance of input descriptors for PC-MG-MoE predictions. Importance values in (c) were obtained by randomly permuting each input descriptor after model training and measuring the resulting increase in output-specific prediction error. Values were normalised within each output and therefore indicate relative model reliance rather than pairwise correlations in the dataset.

### 3.5 Wet-Lab Model–Experiment Comparison

To provide an external model–experiment comparison under laboratory pyrolysis conditions, wet-lab experiments were conducted and compared with PC-MG-MoE outputs. The experimental conditions were used as inputs to the trained model, and the predicted product distributions were compared with the corresponding measured results. The wet-lab data were not used for model training, hyperparameter optimisation, or baseline benchmarking.

The selected validation conditions are summarised in Table S5. As shown in Figure S6, the model–experiment comparison indicates that PC-MG-MoE captured key composition-dependent trends across the validation cases. Pairwise order agreement was used to assess whether the model preserved the relative ranking of validation cases for each output. Aromatics and styrene showed complete pairwise order agreement across the validation cases, while wax, gasoline, solid and BTX also showed high ranking consistency (Table S6). Larger point-wise deviations were observed for some aromatic-related yields, indicating that prediction accuracy remains output-dependent.

Beyond the model–experiment comparison demonstrated in our laboratory, the framework also offers a route for other laboratories to adapt the model using data generated from their own experimental platforms, as discussed in Section 3.6.4.

### 3.6 From Prediction to Model-Guided Experimental Planning and System-Level Capabilities

In plastic pyrolysis research, experimental planning is typically conducted under incomplete information and practical constraints, relying on empirical rules or limited parameter sweeps. To bridge data-driven modelling and experimental decision-making, we deployed the trained PC-MG-MoE model as an interactive web application comprising two main modules, namely a forward “Predictor” that maps user-specified feed composition, temperature and catalyst choice to a full nine-target yield vector, and an “Optimizer” that performs constrained inverse design (Figure 5). In this form, PC-MG-MoE functions not only as a predictive model, but as a queryable reaction-design engine that links fragmented literature evidence to experimentally actionable candidate conditions.

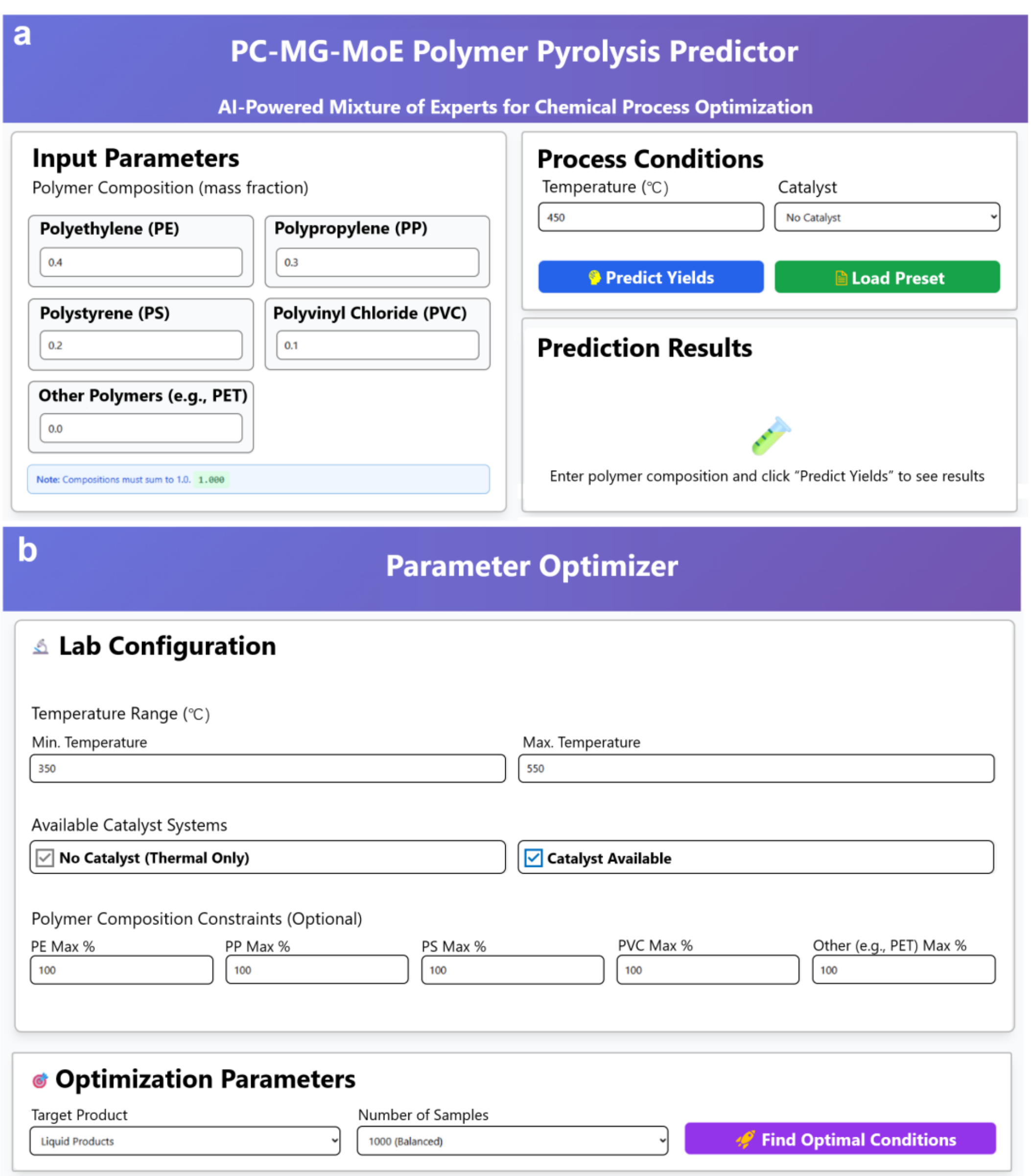

**Figure 5. Web-based deployment of the PC-MG-MoE model for forward prediction and constrained inverse design.** (a) Polymer Pyrolysis Predictor module, enabling forward prediction of pyrolysis product yields under specified reaction conditions; (b) Parameter Optimizer module, enabling constrained inverse design under user-defined compositional and operational constraints. The interfaces are presented in a rearranged form to highlight key functional components.

### 3.6.1 Deployment Framework for Model-Assisted Experimental Planning

The Polymer Pyrolysis Predictor module allows users to input polymer mass fractions, reaction temperature, and catalyst choice. Based on these inputs, the model predicts gas, liquid, solid, gasoline, diesel, wax, aromatics, BTX, and styrene yields, allowing rapid assessment of candidate conditions before committing experimental resources. To ensure physical validity, the Predictor enforces the same mass-balance and non-negativity constraints embedded in PC-MG-MoE, while polymer compositions are required to sum to 100 wt.%. This enables exploratory screening and narrows the feasible operating space before experimentation.

### 3.6.2 Model-Assisted Experimental Decision Support

Beyond forward screening, experimental planning often requires identifying operating conditions that satisfy practical constraints while targeting specific performance objectives. The Parameter Optimizer module addresses this requirement by allowing users to impose experimentally realistic constraints and composition bounds.

Inverse design is formulated as a constrained search over input variables. Polymer fractions are treated as a simplex, with candidate compositions sampled from a Dirichlet distribution subject to user-defined bounds, such as upper limits on PS and PVC. Temperature is sampled uniformly within a specified range, and the catalyst flag can be fixed to reflect laboratory capabilities. For each scenario, 5000 candidates are evaluated and ranked by the target objective after physics projection to enforce feasibility, following the procedure described in the Methods.

By enforcing mass-balance and compositional constraints, the Optimizer excludes physically infeasible regions that could otherwise lead to misleading recommendations. By integrating forward screening with constrained inverse design, the deployed framework moves beyond passive prediction of intuitive trends, such as PS-rich feeds favouring light-oil production [22], towards active experimental decision support under practical constraints.

3.6.3 Case Studies

To illustrate these capabilities, two representative case studies are presented under realistic feedstock and process constraints. Case A maximised aromatics yield under a stringent PS constraint (PS ≤ 10 wt.%), whereas Case B further introduced a product constraint (gas ≤ 30 wt.%), enabling constrained inverse design under both feedstock and product requirements. Figure 6 summarises the top candidate recipes and their sensitivity to perturbations in temperature (±10 °C) and PS content (±2 wt.%).

**Case A.** In this scenario, predicted aromatics were maximised under constraints of PS ≤ 10 wt.%, PVC ≤ 5 wt.%, temperature 400–650 °C and catalyst = 1. The top-ranked candidates clustered near the upper end of the temperature window and achieved predicted aromatics yields of around 17 wt.% (Figure 6(a, c) and Table S7). Despite the low PS content, aromatic selectivity normalised by liquid products reached around 50 wt.%. This was accompanied by elevated gas fractions (~47 wt.%), likely due to extensive cracking and secondary reactions, and moderate liquid yields. This suggests that severe conditions are required to promote aromatics formation from predominantly non-PS feedstocks, consistent with previous observations for low-PS or polyolefin-dominated systems [30]. Sensitivity analysis showed that perturbing temperature or PS content changed the objective by < 0.5 wt.%, indicating a locally stable optimum (Figure S7(a)).

**Case B.** In this scenario, the optimisation problem was extended by adding a product constraint (gas ≤ 30 wt.%), enabling inverse design under both feedstock and product requirements. The PS content was relaxed to PS ≤ 20 wt.% while maintaining PVC ≤ 5 wt.%, temperature 400–650 °C, and catalyst = 1. Under these conditions, the model identified optimal solutions in an intermediate temperature range (~550–580 °C), achieving predicted aromatics yields up to 18.01 wt.% while satisfying the gas ceiling (Figure 6(b, d) and Table S8). This shift reflects a trade-off between aromatisation and over-cracking. While higher temperatures favour aromatics formation, they also promote excessive gas production. By imposing a gas ceiling, the optimiser identifies feasible conditions that balance aromatics yield with an acceptable product distribution. This scenario demonstrates that PC-MG-MoE can perform constrained inverse design under realistic experimental requirements, identifying non-trivial solutions that would be difficult to obtain through conventional trial-and-error. The value of the optimiser is not simply that it identifies the highest predicted aromatics yield, but that it identifies feasible trade-off solutions under simultaneous feedstock, product and process constraints. The predicted optima remained stable under small perturbations (<0.9 wt.% variation), suggesting multiple robust operating points within the feasible region (Figure S7(b)).

Taken together, these case studies demonstrate how PC-MG-MoE supports constraint-aware decision-making by identifying feasible operating conditions that balance feedstock, product and process requirements. The resulting non-trivial optima would be difficult to locate efficiently through conventional trial-and-error or exhaustive parameter sweeps, highlighting the value of model-guided inverse design beyond passive prediction. Although demonstrated for plastic pyrolysis, the same deployment logic could be extended to other thermochemical systems with heterogeneous reporting scopes by redefining the input descriptors and output targets.

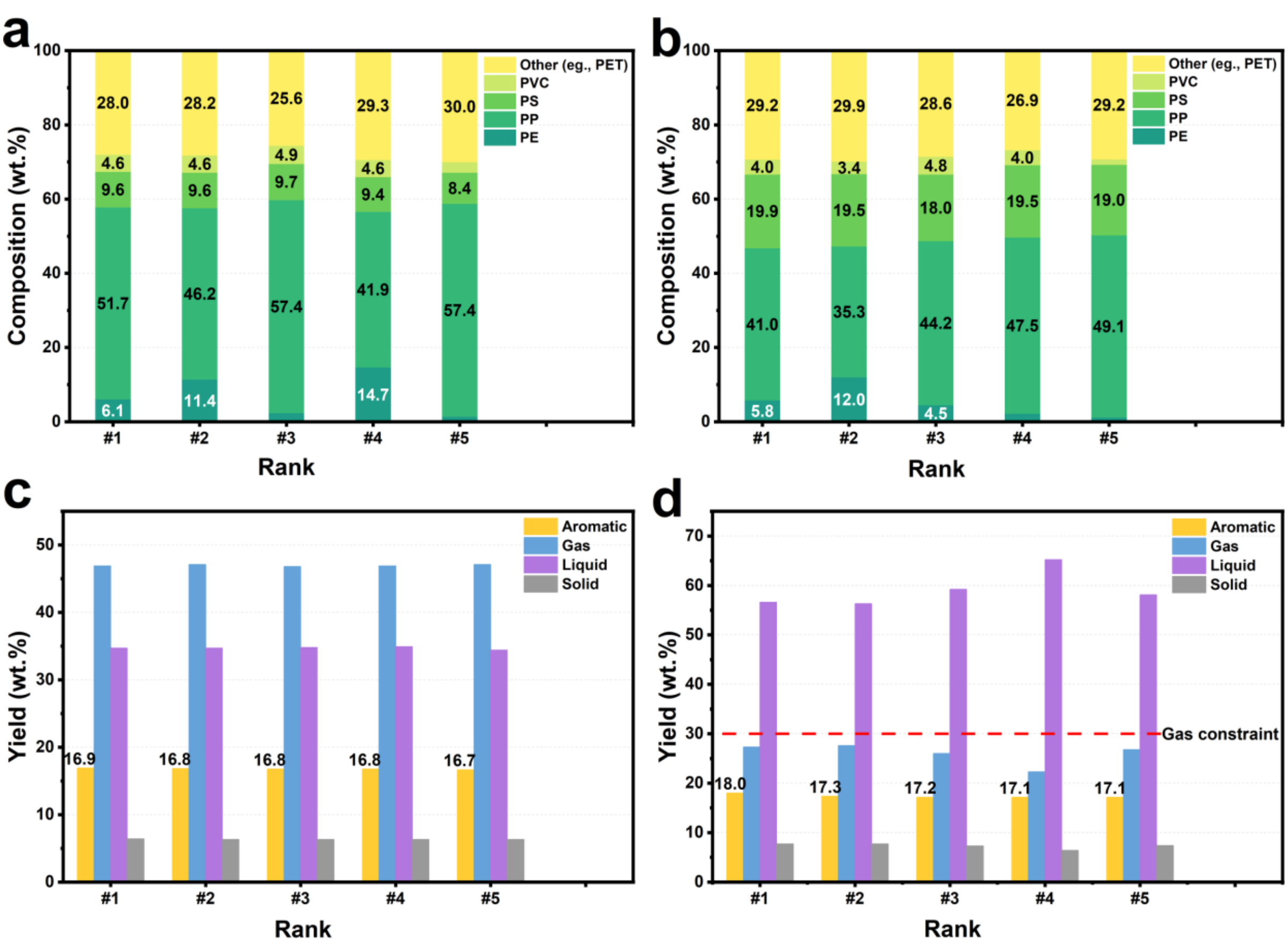

**Figure 6. Model-guided constrained inverse design of feedstock composition and operating conditions for maximising aromatics under realistic constraints.** Panels (a, c) correspond to Case A, and panels (b, d) correspond to Case B.

### 3.6.4 Model-Guided Reduction of Experimental Workload and Laboratory-Specific Adaptation

Beyond identifying feasible optima, the inverse-design workflow supports reduced trial-and-error and experimental workload by pre-screening candidate conditions in silico. For example, a conventional sweep of five representative waste compositions across seven temperatures would require 35 experiments, corresponding to approximately 105 hours of reactor occupancy assuming 3 hours per run. In contrast, the inverse-design module can pre-screen thousands of candidate conditions and nominate a small set of high-value experiments that satisfy practical constraints, such as safety or equipment limits, while targeting specific product objectives. This enables a focused experimental set of approximately 8–10 experiments rather than an exhaustive grid, reducing reactor time to around 27 hours and corresponding to an estimated 75% reduction in both the number of runs and total reactor occupancy. This example illustrates how PC-MG-MoE transforms an exhaustive temperature–composition matrix into a focused, model-guided experimental campaign centred on the most informative and practically relevant operating conditions.

Moreover, the framework can also incorporate additional laboratory-specific data without changing the overall learning architecture. As users generate new data from their own experimental platforms, these data can be progressively incorporated into the framework, allowing the model to retain the general knowledge learned from the broader literature while improving prediction accuracy for specific reactor configurations.

### 3.6.5 System-Level Engineering Capabilities Enabled by PC-MG-MoE

PC-MG-MoE introduces four system-level engineering capabilities that shift reaction-system research from empirical, data-limited exploration toward model-guided knowledge extraction and process design (Figure S8).

First, it reconstructs coherent reaction behaviour from incomplete evidence by enabling physically consistent full-vector inference from partially reported data. Second, it narrows the reaction design space by pre-screening broad temperature–composition regions before experimentation. Third, it enables physics-grounded inverse reasoning under user-defined feedstock, product and process constraints. Fourth, it provides a transferable modelling structure that can be extended to other thermochemical systems with heterogeneous reporting scopes and progressively adapted to laboratory-specific reactor configurations as new experimental data become available. Together, these capabilities transform fragmented experimental records into coherent reaction knowledge, support feasible design-space exploration and provide a route towards adaptable, model-guided experimental planning across complex thermochemical upgrading systems. An extended discussion of these capabilities is provided in Section S6 of the Supporting Information.

## 3.7 Limitations and Outlook

Several limitations should be noted.

First, even after curation, the dataset remains modest (282 experiments). The dataset also inherits substantial heterogeneity in reactor configuration, heating rate, residence time and analytical protocols, many of which are not consistently reported and therefore not included as inputs. This contributes to negative $R^2$ values for several sparsely labelled or source-specific targets under source-grouped validation, particularly for some liquid subfractions and aromatic subfamilies. Target-balanced source-grouped and nested tabular sensitivity analyses reduce the aggregate $R^2$ penalty but do not fully resolve it, indicating that this is a data-and-domain-shift limitation rather than simply a model-tuning issue. Second, the physics projection enforces a restricted set of linear constraints. Richer chemical constraints (for example, element balances, catalyst-specific selectivity limits, or mechanistically informed temperature–selectivity relationships) could further improve plausibility but would require additional metadata. Finally, the inverse-design demonstrations are model-guided screening exercises. The recommended conditions should be treated as hypotheses to be experimentally validated, and uncertainty quantification would be a valuable addition for risk-aware decision-making.

Future work should therefore combine missingness-aware modelling with targeted data acquisition, uncertainty quantification and closed-loop experimental validation, together with more standardised reporting, to progressively densify high-value outputs.

## Conclusion

This work demonstrates that fragmented and partially observed experimental data can be transformed into interpretable, physically feasible and actionable knowledge for plastic upcycling. By explicitly modelling structured missingness as an informative signal, the PC-MG-MoE framework learns directly from heterogeneous, incompletely reported datasets without target imputation, reconstructs physically consistent reaction behaviour, and captures cross-laboratory observation regimes within a unified representation. Model-behaviour analyses further showed that PC-MG-MoE relies on chemically meaningful descriptors rather than acting as a purely black-box predictor, supporting the interpretability of the learned input–output relationships. Together, these elements distinguish PC-MG-MoE from conventional yield-prediction models by framing fragmented and incomplete literature data as a basis for physically constrained, interpretable and experimentally actionable reaction design.

Wet-lab pyrolysis experiments further showed encouraging agreement in key composition-dependent product trends, supporting the experimental relevance of the framework under laboratory conditions. Beyond retrospective analysis, the framework enables forward screening and physics-grounded constrained inverse design under realistic experimental constraints, allowing feasible operating conditions to be identified and prioritised prior to resource-intensive experimentation. In practical scenarios, this capability supports the reduction of exhaustive experimental campaigns into targeted, model-guided exploration while maintaining feasibility. In the illustrative workload analysis, pre-screening reduced the estimated number of experiments and reactor occupancy by approximately 75%. The deployment of the model as an interactive, web-based workflow further translates these capabilities into actionable decision support, bridging the gap between data-driven modelling and experimental implementation. The same deployment framework also provides a route for laboratory-specific adaptation as new data are generated from individual experimental platforms.

More broadly, this work establishes a general paradigm for converting fragmented experimental evidence into actionable reaction knowledge, shifting reaction-system research from data-limited exploration towards model-guided design. While demonstrated for plastic pyrolysis, the proposed framework is potentially extensible to other fragmented thermochemical datasets characterised by heterogeneous reporting and incomplete observations, such as biomass pyrolysis, catalytic upgrading and co-processing systems, highlighting its potential as a transferable and adaptable modelling strategy for thermochemical upgrading systems.

## Code Availability

The source code and trained models developed in this study are publicly available at https://github.com/OliverDOU776/AI4Chem.

## Data Availability

Curated dataset (282 experiments, 73 sources) included in data package.

## CRediT Authorship Contribution Statement

J.B. and Z.W. contributed equally to this work.

## Competing Interests

The authors declare no competing interests.

**Supporting Information**

# From fragmented data to actionable design: Physics-calibrated learning for plastic upcycling

Jingyang Bai [a, †], Zijia Wang [b, †], Xiangyi Long [a], Marcos Millan [c, d], Binjian Nie [e, *], and Mingyue Ding [f, g, *]

[a] Department of Chemical Engineering, Imperial College London, South Kensington Campus, London SW7 2AZ, UK

[b] Department of Electrical and Electronic Engineering, Imperial College London, South Kensington Campus, London SW7 2AZ, UK

[c] Interdisciplinary Research Center for Refining and Advanced Chemicals, King Fahd University of Petroleum and Minerals, Dhahran 31261, Saudi Arabia

[d] Department of Chemical Engineering, King Fahd University of Petroleum & Minerals, Dhahran 31261, Saudi Arabia

[e] Department of Engineering Science, The University of Oxford, Parks Road, OX3 1PJ, Oxford, UK

[f] School of Power and Mechanical Engineering, Wuhan University, Wuhan, 430072, China

[g] Academy of Advanced Interdisciplinary Studies, Wuhan University, Wuhan, 430072, China

[†] These authors contributed equally to this work.

* Corresponding author: binjian.nie@eng.ox.ac.uk; dingmy@whu.edu.cn

**Supporting Information Contents**

This Supporting Information provides detailed mathematical formulations, reactor configuration and analytical methods for wet-lab experiments, extended model-evaluation results, ablation diagnostics, wet-lab model–experiment comparison data, constrained inverse-design outputs and system-level capability analyses supporting the main text.

## S1. Mathematical formulation of Physics-Calibrated, Missingness-Gated and Load-Balanced Mixture-of-Experts (PC-MG-MoE)

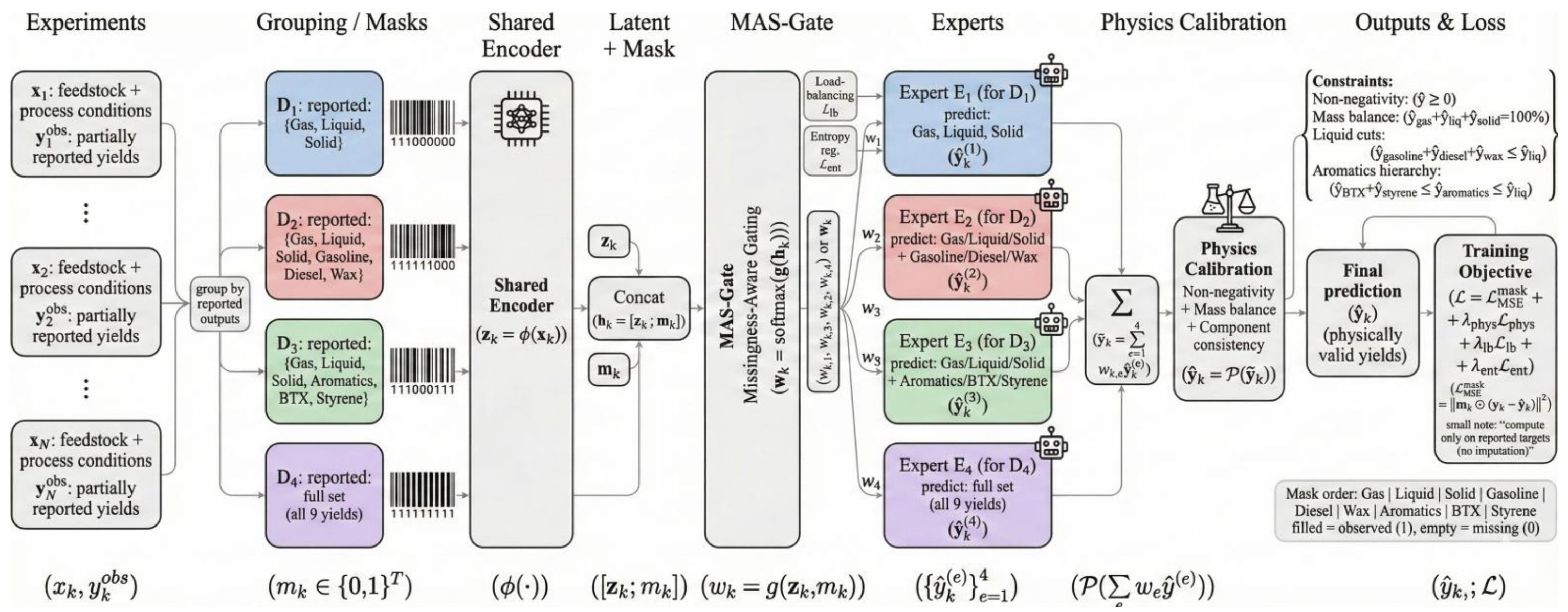

**Figure S1. Physics-calibrated, missingness-gated, and load-balanced mixture-of-experts (PC-MG-MoE) architecture for learning from heterogeneous and incompletely reported pyrolysis datasets.** Literature-derived experiments are represented by process descriptors, partially reported yield vectors and binary missingness masks. The missingness-aware gate assigns samples to regime-specialised experts according to both process conditions and reporting scope. Expert outputs are aggregated and passed through a physics-calibration projection operator to enforce non-negativity, mass balance and hierarchical consistency among product groups. Training is performed using masked supervision on reported targets without target imputation, together with regularisation terms that stabilise expert utilisation.

This section provides the full mathematical formulation of the PC-MG-MoE architecture described in Section 2.2, including the shared encoder, observation-mask concatenation, Missingness-Aware Softmax Gating (MAS-Gate), regime-specialised experts, physics calibration, masked regression loss and regularisation terms.

As shown in Figure S1, PC-MG-MoE integrates (i) a shared representation, (ii) missingness-aware routing, (iii) regime-specialised experts, and (iv) physics calibration.

**Shared encoder.** A shared encoder $\phi(\cdot)$ maps each input $x_k$ to a latent representation:

$$z_k = \phi(x_k).$$

**Latent–mask concatenation.** To make routing explicitly aware of the reporting pattern, we concatenate the latent representation with the mask:

$$h_k = [z_k;\ m_k].$$

**Missingness-Aware Softmax Gating (MAS-Gate).** The gating network $g(\cdot)$outputs a probability simplex over $E = 4$ experts:

$$w_k = \text{softmax}\big(g(h_k)\big), \qquad w_k = \big(w_{k,1}, w_{k,2}, w_{k,3}, w_{k,4}\big),$$

$$\sum_{e=1}^{4} w_{k,e} = 1, w_{k,e} \geq 0.$$

By conditioning on both $z_k$ and $m_k$, MAS-Gate can route each experiment toward experts compatible with its output structure, while still allowing feature-dependent routing within the same reporting regime.

**Inference-time protocol.** At inference, the observation mask m reflects the user's intended prediction scope. For full nine-target prediction (the default), m = 1 is used. The mask is not derived from unknown future observations but specified by the user.

**Regime-specialised experts.** We instantiate four experts $\{E_e\}_{e=1}^{4}$, each designed to specialise on one regime:

$$\hat{y}_k^{(e)} = E_e(z_k), e \in \{1,2,3,4\}.$$

Expert $E_1$ targets {Gas, Liquid, Solid}, $E_2$ targets {Gas, Liquid, Solid, Gasoline, Diesel, Wax}, $E_3$ targets {Gas, Liquid, Solid, Aromatics, BTX, Styrene}, and $E_4$ targets the full set. In implementation, each expert can output a $T$-dimensional vector $\hat{y}_k^{(e)}$ while the supervised loss is masked to the targets relevant for the sample (and, optionally, for the expert's regime), ensuring that experts learn from all compatible partial labels without requiring any label completion.

**Mixture aggregation.** Expert predictions are combined with MAS-Gate weights:

$$\tilde{y}_k = \sum_{e=1}^{4} w_{k,e} \, \hat{y}_k^{(e)}.$$

**Physics calibration**

Direct regression on heterogeneous, partially observed yields can produce chemically invalid outputs (e.g., negative yields or mass imbalance). Therefore, a physics-calibration operator $\mathcal{P}(\cdot)$ is introduced that maps the mixed prediction $\tilde{y}_k$ to a physically consistent final prediction:

$$\hat{y}_k = \mathcal{P}(\tilde{y}_k).$$

The feasible set enforced by $\mathcal{P}$ follows the constraints shown in Figure S1:

1. **Non-negativity:** $\hat{y} \geq 0$.

2. **Overall mass balance (macro products):**

$$\hat{y}_{gas} + \hat{y}_{liq} + \hat{y}_{solid} = 100\%.$$

3. **Consistency of liquid cuts (when applicable):**

$$\hat{y}_{gasoline} + \hat{y}_{diesel} + \hat{y}_{wax} \leq \hat{y}_{liq}.$$

4. **Aromatics hierarchy (when applicable):**

$$\hat{y}_{BTX} + \hat{y}_{styrene} \leq \hat{y}_{aromatics} \leq \hat{y}_{liq}.$$

In this work, $\mathcal{P}$ is implemented as a projection operator onto the feasible constraint set defined above. This projection calibrates the mixed prediction $\tilde{y}_k$ to satisfy the imposed non-negativity, macro-product mass-balance, liquid-cut consistency and aromatic-hierarchy constraints. In this way, physics calibration ensures that the final yield vector follows the predefined chemical mass-balance and compositional-hierarchy logic.

PC-MG-MoE is trained end-to-end with a composite objective comprising masked regression, physics, load-balancing, and entropy terms:

$$\mathcal{L} = \mathcal{L}_{MSE}^{mask} + \lambda_{phys}\mathcal{L}_{phys} + \lambda_{lb}\mathcal{L}_{lb} + \lambda_{ent}\mathcal{L}_{ent}.$$

The weighting coefficients were selected through preliminary hyperparameter screening and then kept fixed across all reported experiments.

**Masked regression loss (no imputation).** Supervision is computed only on reported targets using $m_k$:

$$\mathcal{L}_{MSE}^{mask} = \frac{1}{N}\sum_{k=1}^{N} \frac{1}{\| m_k \|_1} \| m_k \odot (y_k^{obs} - \hat{y}_k) \|_2^2,$$

where $\odot$ denotes elementwise multiplication. This formulation uses all partially reported experiments directly, without discarding samples or imputing missing targets.

**Physics loss.** We include a physics term $\mathcal{L}_{phys}$ to encourage predictions to lie near the physically feasible set and to stabilise optimisation. A natural choice consistent with the calibration operator is the calibration magnitude:

$$\mathcal{L}_{phys} = \frac{1}{N}\sum_{k=1}^{N} \| \tilde{y}_k - \mathcal{P}(\tilde{y}_k) \|_2^2,$$

or, equivalently, a sum of hinge penalties measuring violations of the constraints before calibration.

**Load balancing and entropy regularisation.** To prevent routing collapse and promote meaningful expert specialisation under heterogeneous measurement scopes, we regularise the gate. Let $\bar{w}_e = \frac{1}{N}\sum_{k=1}^{N} w_{k,e}$ denote the average utilisation of expert $e$. We define a load-balancing term

$$\mathcal{L}_{lb} = \sum_{e=1}^{4} (\bar{w}_e - \frac{1}{4})^2,$$

which discourages degenerate solutions where only a subset of experts receives all probability mass. In addition, we apply an entropy regulariser

$$\mathcal{L}_{ent} = -\frac{1}{N}\sum_{k=1}^{N} H(w_k) = \frac{1}{N}\sum_{k=1}^{N}\sum_{e=1}^{4} w_{k,e} \log w_{k,e},$$

so that minimising $\mathcal{L}_{ent}$ encourages higher-entropy routing distributions where appropriate, improving optimisation stability and enabling diverse expert activation.

By combining missingness-aware routing ($m_k$ enters MAS-Gate), masked supervision (loss computed only on observed targets), physics calibration ($\hat{y}_k = \mathcal{P}(\tilde{y}_k)$), and balanced gating regularisation ($\mathcal{L}_{lb}, \mathcal{L}_{ent}$), PC-MG-MoE learns from heterogeneous and pervasively incomplete pyrolysis datasets while producing chemically consistent product-yield distributions.

## S2. Reactor Configuration and Analytical Methods for Wet-Lab Experiments

The wet experiments were conducted in a plastic pyrolysis reactor. The reactor flow sheet is shown in Figure S2. Liquid products were analysed qualitatively and quantitatively using gas chromatography–mass spectrometry/flame ionisation detection (GC–MS/FID, Shimadzu GCMS-QP2020NX) equipped with an SH-Rxi-5ms capillary column (L = 30 m, ID = 0.25 mm, df = 0.25 μm). Solid products were collected inside the reactor after the experiments, and gas yield was calculated by difference.

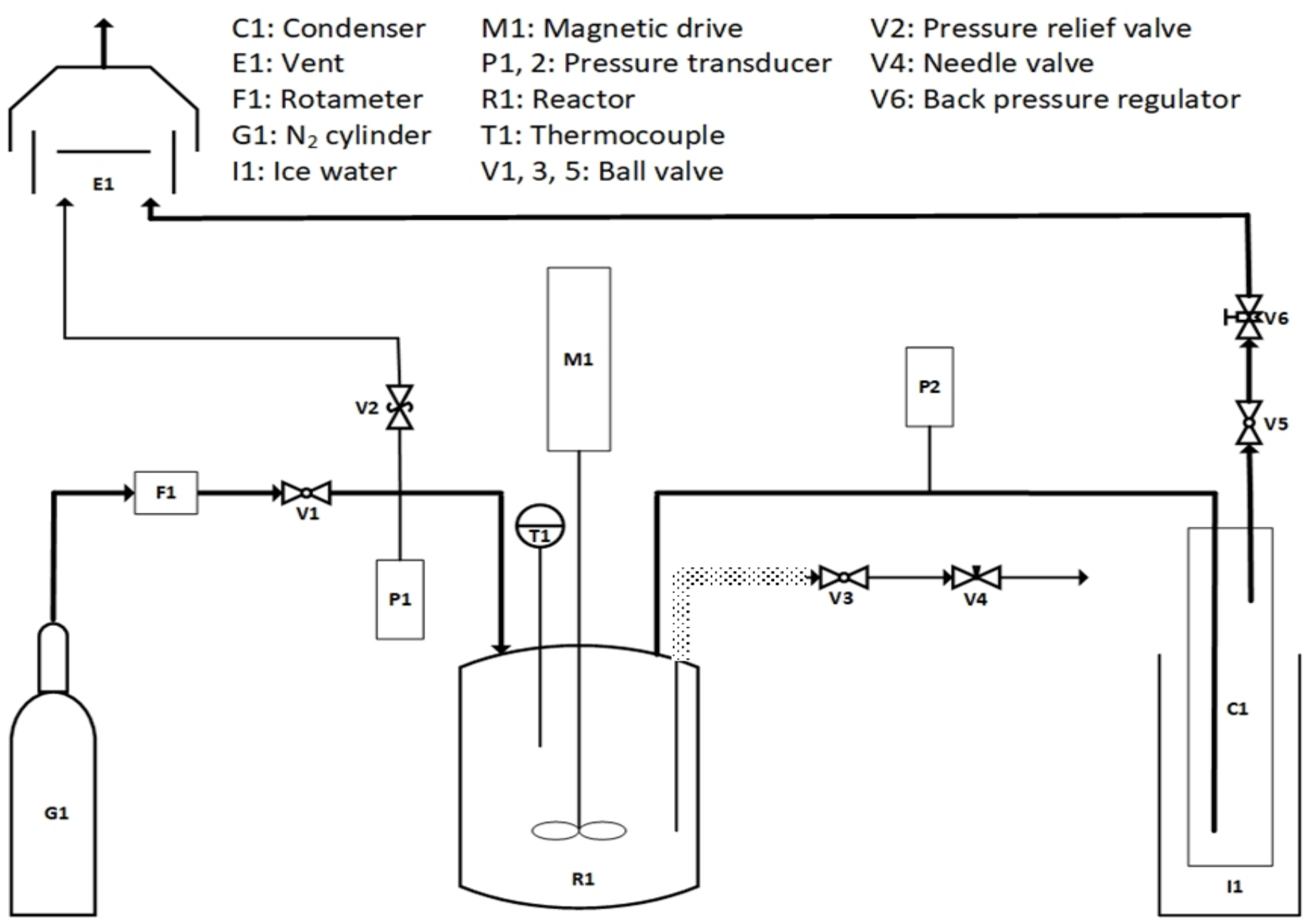

**Figure S2. Flow sheet of the laboratory-scale pyrolysis reactor used for wet-lab experiments.**

## S3. Structured Reporting Regimes and Supplementary Model Evaluation

**Table S1.** Recurring reporting regimes (binary masks)

| Regime ID | Mask | Count | Observation pattern |
|---|---|---|---|
| 4 | 1,1,1,0,0,0,0,0,0 | 95 | gas; liquid; solid |
| 7 | 1,1,1,0,0,0,1,1,1 | 63 | gas; liquid; solid; aromatics; BTX; styrene |
| 6 | 1,1,1,0,0,0,1,0,0 | 35 | gas; liquid; solid; aromatics |
| 10 | 1,1,1,1,1,1,1,1,1 | 31 | gas; liquid; solid; gasoline; diesel; wax; aromatics; BTX; styrene |
| 3 | 0,1,0,1,1,1,1,0,0 | 21 | liquid; gasoline; diesel; wax; aromatics |
| 8 | 1,1,1,1,1,1,0,0,0 | 20 | gas; liquid; solid; gasoline; diesel; wax |
| 9 | 1,1,1,1,1,1,1,0,0 | 14 | gas; liquid; solid; gasoline; diesel; wax; aromatics |
| 5 | 1,1,1,0,0,0,0,1,1 | 1 | gas; liquid; solid; BTX; styrene |
| 1 | 0,1,0,0,0,0,1,1,1 | 1 | liquid; aromatics; BTX; styrene |
| 2 | 0,1,0,1,1,1,0,0,0 | 1 | liquid; gasoline; diesel; wax |

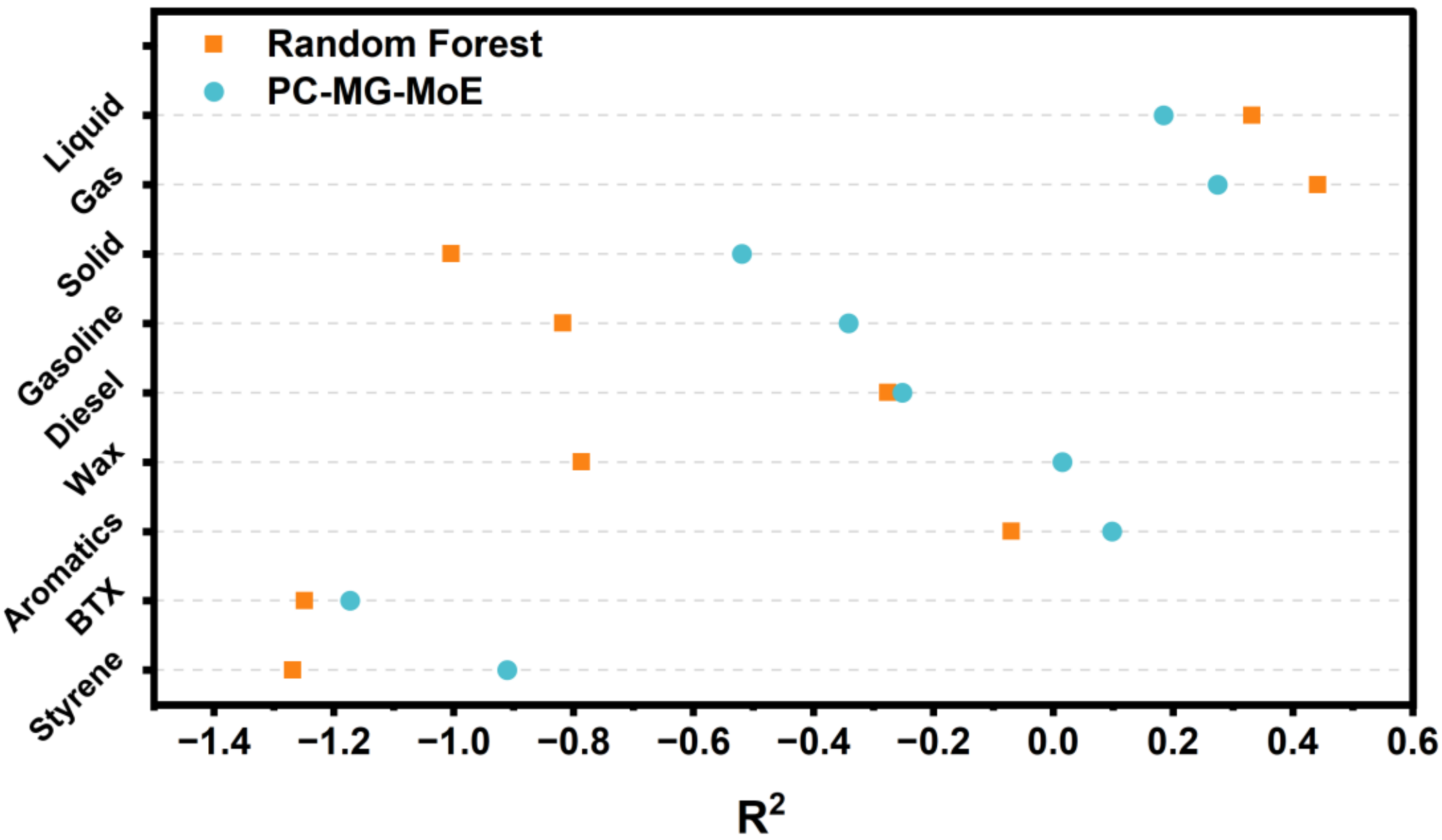

**Figure S3. Baseline comparison under source-grouped evaluation.** Per-target $R^2$, comparing PC-MG-MoE and Random Forest across targets, illustrates both the recoverable macro-yield signals and the limited cross-source variance-explained performance for sparse high-value outputs.

**Table S2.** PC-MG-MoE per-target metrics on the source-grouped split.

| Target | MSE ↓ | MAE ↓ | $R^2$ ↑ |
|---|---|---|---|
| Gas | 436 ± 146 | 16.0 ± 3.8 | 0.27 ± 0.44 |
| Liquid | 655 ± 276 | 20.5 ± 3.6 | 0.18 ± 0.26 |
| Solid | 343 ± 348 | 9.72 ± 5.0 | −0.52 ± 0.55 |
| Gasoline | 444 ± 269 | 16.9 ± 6.1 | −0.34 ± 0.91 |
| Diesel | 132 ± 78 | 9.37 ± 3.4 | −0.25 ± 0.33 |
| Wax | 598 ± 477 | 17.0 ± 8.2 | 0.02 ± 0.42 |
| Aromatics | 620 ± 492 | 18.3 ± 8.3 | 0.10 ± 0.57 |
| BTX | 111 ± 79 | 7.7 ± 2.0 | −1.17 ± 0.39 |
| Styrene | 413 ± 766 | 10.4 ± 14.0 | −0.91 ± 1.67 |

Sensitivity analysis using a target-balanced source-grouped split reduced aggregate PC-MG-MoE $R^2$ from −0.292 to −0.115 while preserving zero source-ID and experiment-signature overlap. In this sensitivity split, gas ($R^2$ = 0.481), liquid (0.188), aromatics (0.225) and styrene (0.049) were positive on average, whereas gasoline (−0.923), solid (−0.425), diesel (−0.287), BTX (−0.197) and wax (−0.142) remained limited. An independent nested tabular sweep on the same split gave a similar aggregate $R^2$ of −0.111 and MAE of 13.80 wt.%, with BTX becoming weakly positive ($R^2$ = 0.071) but gasoline remaining strongly negative ($R^2$ = −1.462). These analyses indicate that fold imbalance contributes to the most severe negative $R^2$ values, but sparse high-value product prediction remains only partially recoverable from the current metadata.

**Table S3.** Source-grouped component ablations for PC-MG-MoE. Values are reported as mean ± standard deviation over five source-held-out folds. These ablations assess the contribution of mask-aware gating and routing strategy under the same cross-source evaluation protocol used for baseline comparison.

| Model | MSE↓ | MAE↓ | $R^2$↑ |
|---|---|---|---|
| Full pure PC-MG-MoE | 454.8 ± 174.8 | 14.68 ± 3.55 | −0.56 ± 0.48 |
| No mask-aware gate | 456.7 ± 205.1 | 14.65 ± 3.31 | −0.50 ± 0.57 |
| **Hard routing** | **501.0 ± 188.0** | **15.53 ± 3.42** | **−0.86 ± 1.05** |

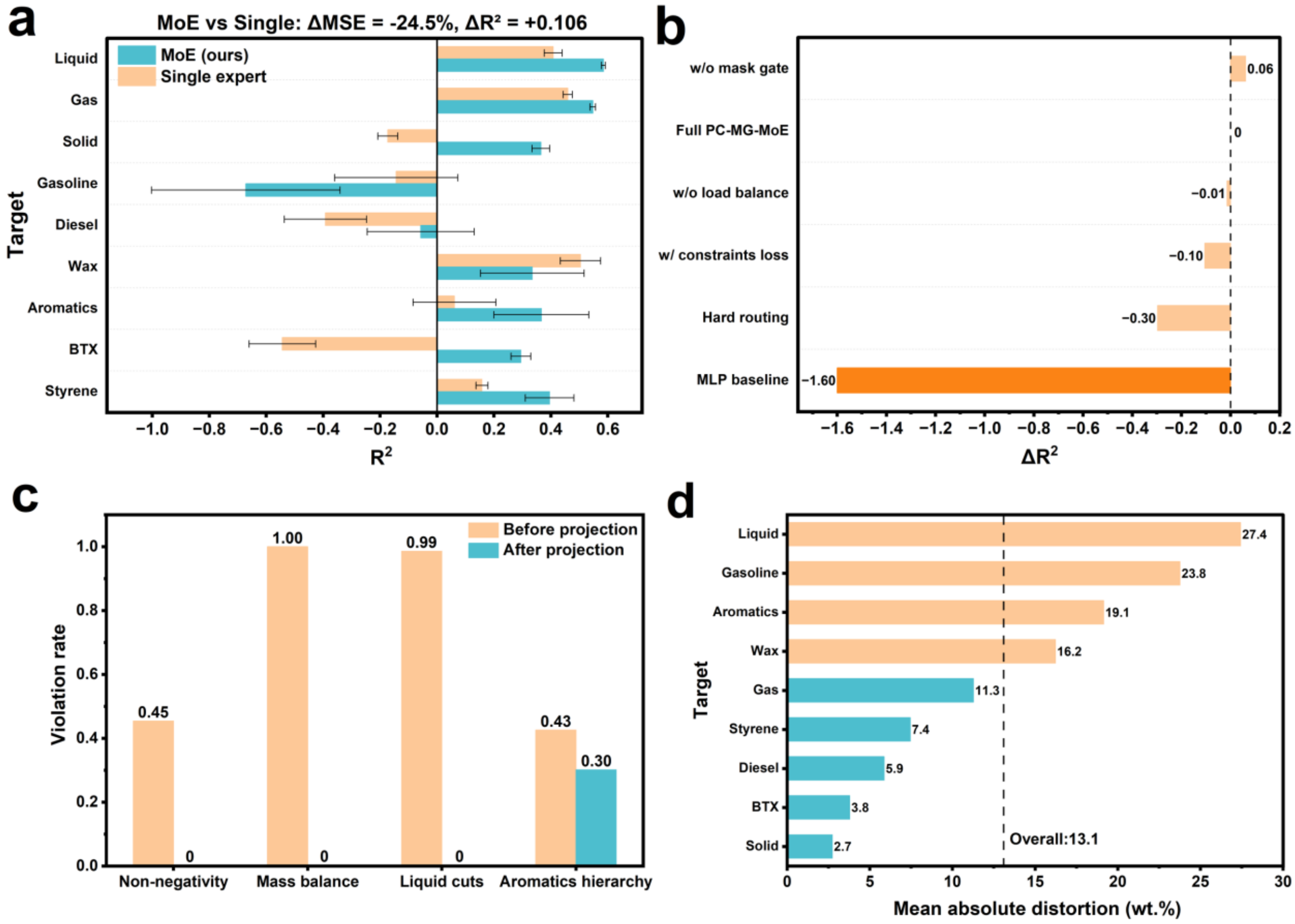

**Figure S4. Ablation studies and physics calibration of the PC-MG-MoE model.** (a) Per-output $R^2$ comparing single-expert and MoE models under five-fold cross-validation, assessing representational capacity and not used as cross-laboratory generalisation metrics; (b) Component ablation showing $\Delta R^2$ relative to the full PC-MG-MoE model under the source-grouped evaluation protocol; (c) Violation rates before and after projection; (d) Calibration distortion by target. Error bars indicate variability across source-grouped cross-validation folds where applicable.

**Table S4.** Within-source architecture diagnostic comparing single-expert and MoE models. These values are not source-grouped generalisation results and are therefore not directly comparable with Table 1 or Table S2.

| Target | Single-expert | | MoE | |
|---|---|---|---|---|
| | Mean $R^2$ | SD [1] | Mean $R^2$ | SD |
| Liquid | 0.4084 | 0.0315 | 0.5848 | 0.0069 |
| Gas | 0.4595 | 0.0160 | 0.5475 | 0.0096 |
| Solid | -0.1729 | 0.0349 | 0.3649 | 0.0310 |
| Gasoline | -0.1433 | 0.2167 | -0.6716 | 0.3315 |
| Diesel | -0.3921 | 0.1445 | -0.0573 | 0.1881 |
| Wax | 0.5042 | 0.0708 | 0.3348 | 0.1823 |
| Aromatics | 0.0614 | 0.1456 | 0.3668 | 0.1668 |
| BTX | -0.5433 | 0.1172 | 0.2947 | 0.0345 |
| Styrene | 0.1579 | 0.0210 | 0.3956 | 0.0857 |

[1] SD stands for standard deviation.

Because Table S4 uses a non-source-grouped diagnostic split, it is retained only to show that the MoE architecture can increase representational capacity when laboratory/source-level domain shift is not held out. Accordingly, the source-grouped ablation results in the main text should be used for generalisation claims.

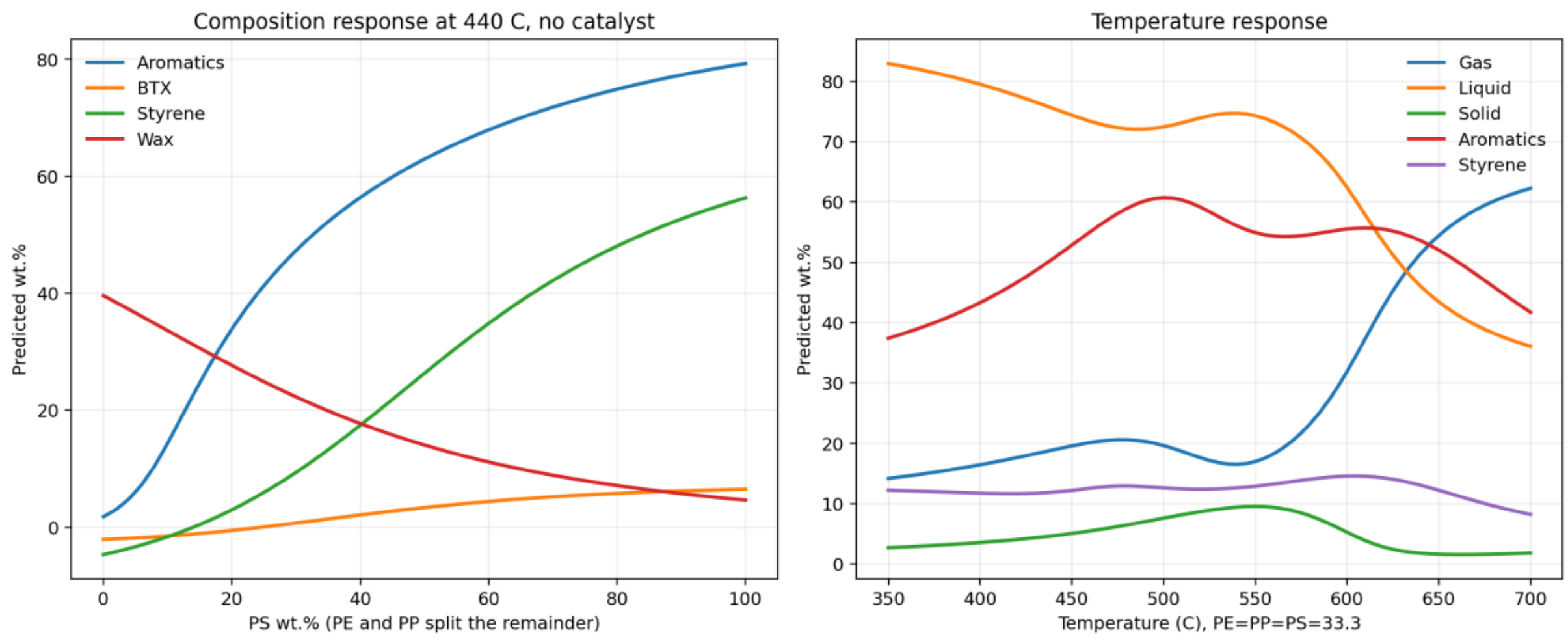

**Figure S5. Controlled composition and temperature responses predicted by PC-MG-MoE.** Controlled-response analysis was performed by varying one input factor while keeping the remaining conditions fixed. The composition response was evaluated by increasing PS content at 440 °C without catalyst, while PE and PP shared the remaining fraction. The temperature response was evaluated at PE = PP = PS = 33.3 wt.% without catalyst.

## S4. Wet-Lab Model–Experiment Comparison

**Table S5.** Wet-lab validation conditions used for comparison with PC-MG-MoE predictions.

| Case | PE<br>wt.% | PP<br>wt.% | PS<br>wt.% | Temperature<br>℃ |
|---|---|---|---|---|
| R1 | 80 | 10 | 10 | 440 |
| R2 | 60 | 20 | 20 | 440 |
| R3 | 33.3 | 33.3 | 33.3 | 440 |
| R4 | 30 | 60 | 10 | 440 |
| R5 | 20 | 60 | 20 | 440 |
| R6 | 20 | 20 | 60 | 440 |

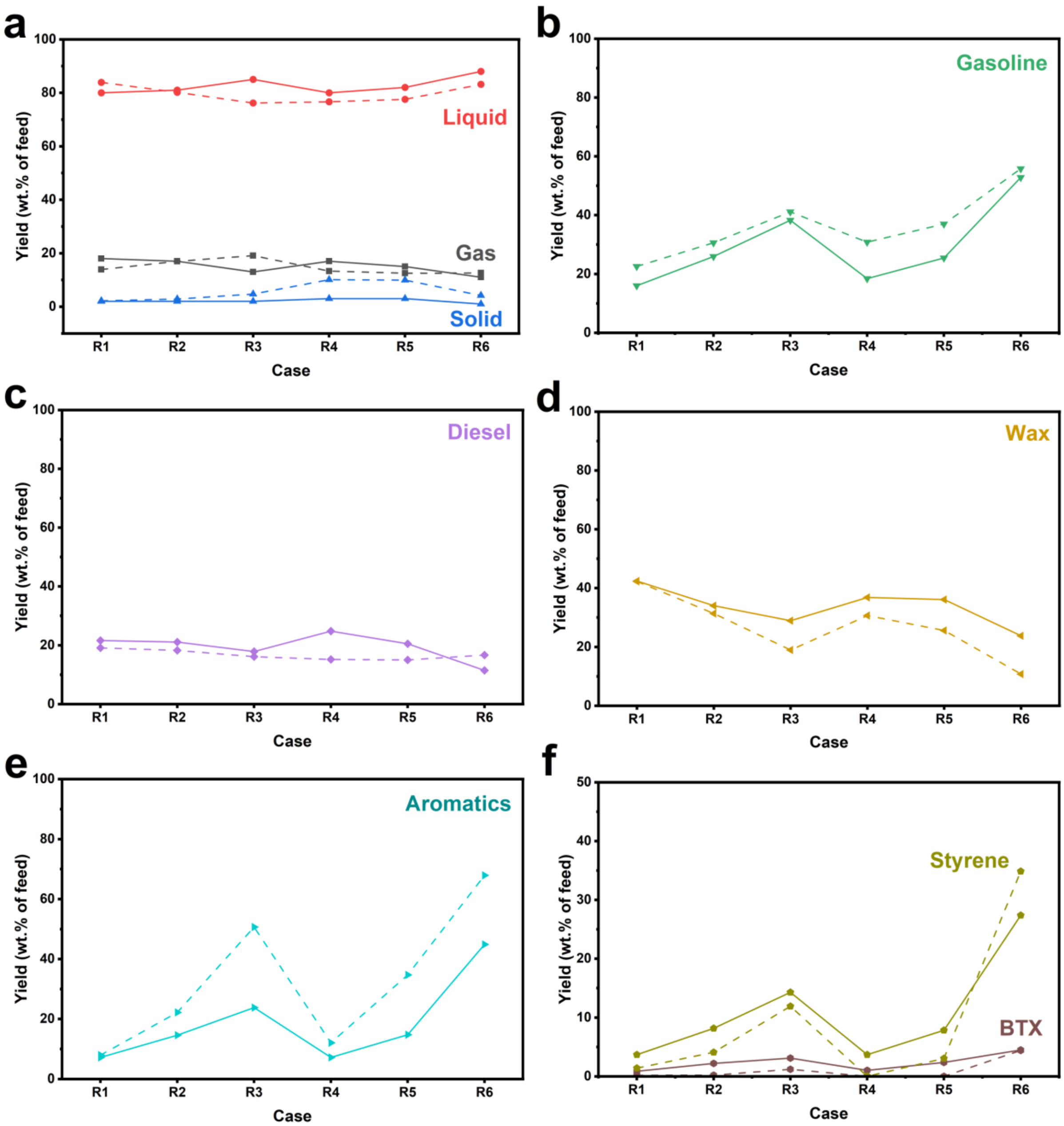

**Figure S6. Case-wise comparison between wet-lab measurements and PC-MG-MoE predictions**. Solid lines denote wet-lab measurements, and dashed lines denote PC-MG-MoE predictions. The plots illustrate case-to-case variation trends.

**Table S6.** Trend-agreement metrics for the wet-lab model–experiment comparison.

| Output | Spearman's ρ | Kendall's τ | Pairwise order agreement | MAE | RMSE |
|---|---|---|---|---|---|
| Gas | 0.20 | 0.14 | 57% | 3.01 | 3.54 |
| Liquid | −0.14 | −0.14 | 43% | 4.59 | 5.25 |
| Solid | 0.68 | 0.54 | 82% | 3.48 | 4.39 |
| Gasoline | 0.83 | 0.73 | 87% | 8.98 | 9.89 |
| Diesel | 0.14 | 0.20 | 60% | 3.86 | 4.69 |
| Wax | 0.94 | 0.87 | 93% | 5.80 | 7.23 |
| Aromatics | 0.99 | 0.97 | 100% | 13.85 | 16.99 |
| BTX | 0.71 | 0.60 | 80% | 1.70 | 1.97 |
| Styrene | 0.99 | 0.97 | 100% | 4.38 | 4.72 |

## S5. Constrained Inverse Design and Sensitivity Analysis

**Table S7.** Top-ranked candidate operating conditions identified by the PC-MG-MoE model under constrained inverse design (Case A).

| Rank | PE | PP | PS | PVC | Other (eg., PET) | Temp. (°C) | **Aromatics** | Gas | Liquid | Solid |
|---|---|---|---|---|---|---|---|---|---|---|
| 1 | 6.1 | 51.7 | 9.6 | 4.6 | 28.0 | 649 | **16.90** | 46.9 | 34.7 | 6.4 |
| 2 | 11.4 | 46.2 | 9.6 | 4.6 | 28.2 | 650 | **16.83** | 47.1 | 34.7 | 6.3 |
| 3 | 2.4 | 57.4 | 9.7 | 4.9 | 25.6 | 648 | **16.76** | 46.8 | 34.8 | 6.3 |
| 4 | 14.7 | 41.9 | 9.4 | 4.6 | 29.3 | 649 | **16.75** | 46.9 | 34.9 | 6.3 |
| 5 | 1.4 | 57.4 | 8.4 | 2.8 | 30.0 | 648 | **16.65** | 47.1 | 34.4 | 6.3 |

All compositions and yields are reported in wt.% of feed.

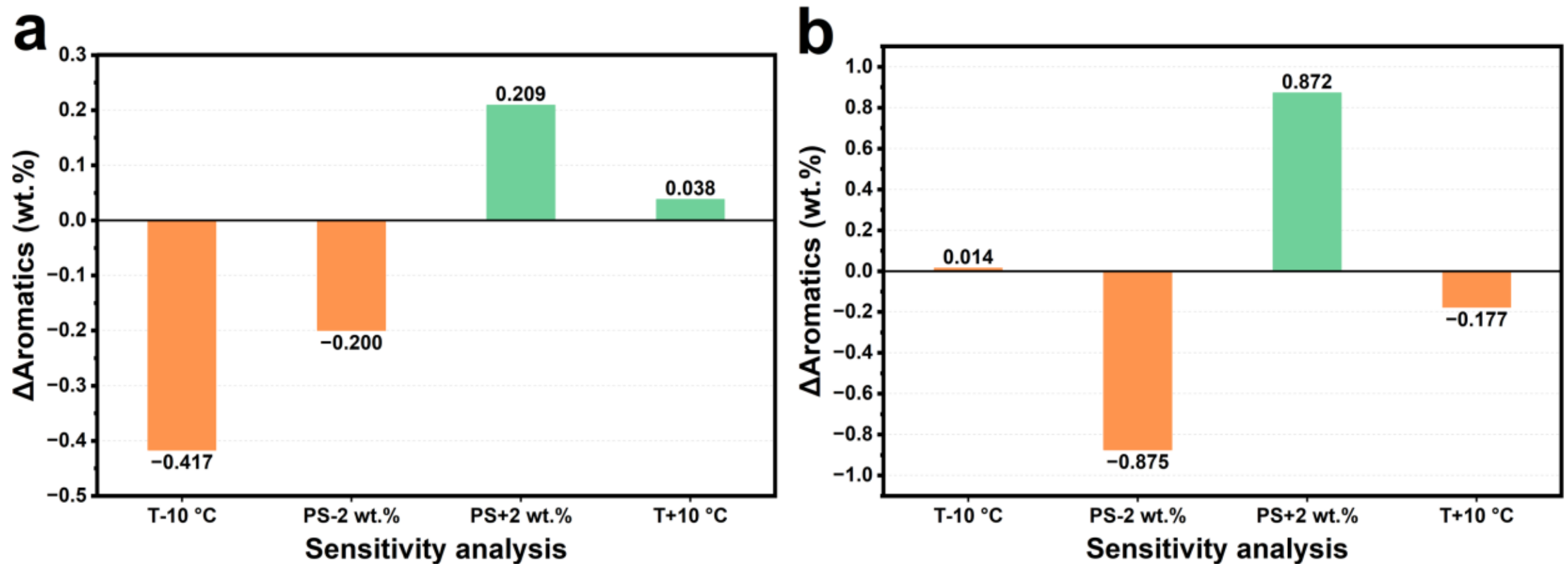

**Figure S7. Sensitivity of predicted aromatics yield to perturbations in temperature and PS content** for (a) Case A and (b) Case B.

**Table S8.** Top-ranked candidate operating conditions identified by the PC-MG-MoE model under constrained inverse design (Case B).

| Rank | PE | PP | PS | PVC | Other (eg., PET) | Temp. (°C) | **Aromatics** | Gas | Liquid | Solid |
|---|---|---|---|---|---|---|---|---|---|---|
| 1 | 5.8 | 41.0 | 19.9 | 4.0 | 29.2 | 580 | **18.01** | 27.3 | 56.6 | 7.7 |
| 2 | 12.0 | 35.3 | 19.5 | 3.4 | 29.9 | 579 | **17.34** | 27.6 | 56.3 | 7.7 |
| 3 | 4.5 | 44.2 | 18.0 | 4.8 | 28.6 | 570 | **17.16** | 26.0 | 59.2 | 7.3 |
| 4 | 2.2 | 47.5 | 19.5 | 4.0 | 26.9 | 550 | **17.14** | 22.3 | 65.2 | 6.4 |
| 5 | 1.2 | 49.1 | 19.0 | 1.5 | 29.2 | 574 | **17.13** | 26.8 | 58.1 | 7.4 |

All compositions and yields are reported in wt.% of feed.

## S6. Extended Discussion of System-Level Engineering Capabilities Enabled by PC-MG-MoE

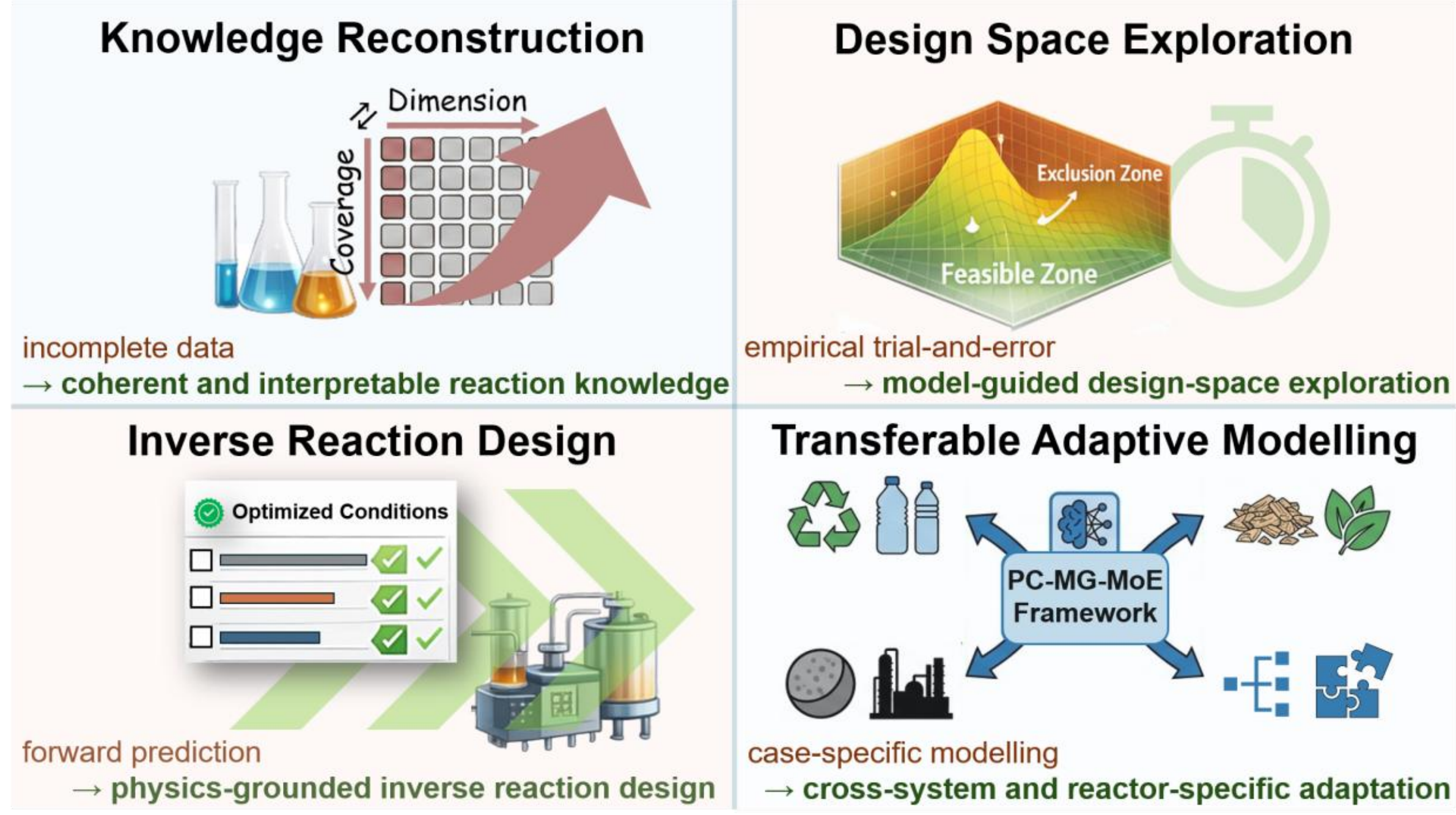

**Figure S8. System-level engineering capabilities enabled by PC-MG-MoE.** The framework transforms incomplete literature data into coherent and interpretable reaction knowledge, supports feasible design-space exploration and physics-grounded inverse design, and provides a transferable and adaptable modelling structure for heterogeneous thermochemical systems and laboratory-specific reactor configurations.

PC-MG-MoE introduces four system-level engineering capabilities that shift reaction-system research from empirical, data-limited exploration toward model-guided knowledge extraction and process design (Figure S8). These capabilities arise from the model's combination of physics-guided calibration, missingness-aware learning, multi-output co-modelling and interpretable model-behaviour diagnostics.

First, PC-MG-MoE reconstructs coherent reaction behaviour from incomplete evidence. Using mask-aware learning, it retains all 282 curated experiments rather than relying on complete-case filtering or target imputation, despite only around 10% of entries reporting all outputs. This enables physically consistent full-vector inference, where experiments reporting only macro yields can still be associated with plausible gasoline/diesel/wax distributions and aromatic subfamilies. The same structure also allows users to incorporate laboratory-specific data without redefining the model architecture, improving adaptation to their own reactor and measurement settings.

Second, the framework supports model-guided exploration and narrowing of the reaction design space. By rapidly evaluating broad temperature–composition regions before experimentation, PC-MG-MoE can identify high-value regions, uninformative zones and physically implausible combinations in advance. This shifts experimental planning from exhaustive one-factor-at-a-time sweeps towards targeted, information-efficient design, allowing laboratory effort to be focused on the most informative operating conditions.

Third, PC-MG-MoE enables physics-grounded inverse reasoning for feasible reaction design. By embedding non-negativity, mass-balance, compositional consistency and hierarchical reaction logic, the model constrains recommendations to physically admissible regions. Users can specify feedstock limits, equipment constraints and product objectives, after which the optimizer ranks feasible candidate conditions and provides simple sensitivity checks. This moves the model beyond passive prediction towards reaction design under explicit physical and practical constraints.

Fourth, the framework provides a transferable and adaptable structure for other thermochemical systems with heterogeneous or incomplete reporting, such as biomass pyrolysis, catalytic upgrading and co-processing pathways. Because PC-MG-MoE operates at the level of data structure and physical constraints rather than plastic-specific chemistry alone, new reporting regimes can be incorporated through new observation masks or expert adaptation. The same structure can also support laboratory-specific adaptation as users generate new data from their own experimental platforms. These data can be progressively incorporated to improve prediction accuracy for specific reactor configurations while retaining the general knowledge learned from broader literature. As reporting becomes more standardised, the same framework can be retrained to improve sparse targets and include additional descriptors such as heating rate, residence time and catalyst identity.

Together, these capabilities transform fragmented experimental records into coherent and interpretable reaction knowledge, narrow the reaction design space before experimentation, enable physics-grounded inverse design and provide a route towards transferable and laboratory-adaptable modelling across complex thermochemical upgrading systems.